\documentclass{article}

\usepackage{microtype}
\usepackage{graphicx}
\usepackage{subcaption}
\usepackage{booktabs} %
\usepackage{adjustbox}
\usepackage{siunitx}

\usepackage{needspace}

\usepackage{hyperref}

\usepackage[accepted]{icml2026}

\usepackage{pgfplots}
\pgfplotsset{compat=1.18}

\usepackage{amsmath}
\usepackage{amssymb}
\usepackage{mathtools}
\usepackage{amsthm}

\usepackage[capitalize,noabbrev]{cleveref}

\usepackage{times}
\usepackage{latexsym}
\usepackage{enumitem}

\usepackage[T1]{fontenc}
\usepackage[utf8]{inputenc}

\usepackage{microtype}

\usepackage{inconsolata}
\usepackage[scaled=0.85]{helvet}          %

\usepackage{graphicx}
\usepackage{xspace}
\usepackage{csquotes}
\makeatletter
\xspaceaddexceptions{\csq@qclose@i}
\makeatletter
\usepackage{amsfonts}
\usepackage{bbm}
\usepackage{cancel}
\usepackage{caption}
\usepackage{subcaption}
\usepackage{arydshln}

\usepackage{pgfplots}
\pgfplotsset{compat=1.18}

    \crefname{section}{\S}{\S\S}
    \crefformat{section}{\S#2#1#3}
    \crefname{figure}{Fig.}{Fig.}
    \crefname{algorithm}{Alg.}{Alg.}
    \crefname{line}{line}{lines}
    \crefname{appendix}{App.}{}
    \crefname{equation}{eq.}{eqs.}
    \crefformat{equation}{eq.~#2#1#3}
    \crefrangeformat{equation}{eqs.~#3#1#4 to #5#2#6}
    \crefmultiformat{equation}{eqs.~#2#1#3}{ and~#2#1#3}{, #2#1#3}{ and~#2#1#3}
    \crefname{table}{Table}{Tables}
    \crefname{proposition}{Proposition}{Propositions}
    \crefname{assumption}{Assump.}{Assumps.}
    \crefname{definition}{Definition}{Definitions}
    \crefname{theorem}{Thm.}{Thm.}

\usepackage{etoolbox} %
\makeatletter
\DeclareRobustCommand*{\escapeus}[1]{%
  \begingroup
    \edef\x{\endgroup
      \noexpand\@escapeusaux{\unexpanded{#1}}}%
  \x}
\def\@escapeusaux#1{%
  \begingroup
    \@activeus
    \scantokens{#1\endinput}%
  \endgroup}
\begingroup\lccode`\~=`\_
\lowercase{\endgroup
  \def\@activeus{\catcode`\_=\active \let~\_}}
\makeatother

\newcommand{\myemph}[1]{\textsf{{\escapeus{#1}}}}

\newcommand{\eg}{\emph{e.g.}\@\xspace}
\newcommand{\ie}{\emph{i.e.}\@\xspace}
\newcommand{\ngram}{$n$-gram\xspace}

\newcommand{\R}{\mathbb{R}}
\newcommand{\chspan}[1]{\text{\rmfamily\textit{``#1''}}\xspace}

\newcommand{\tokenprint}[1]{\text{`#1'}}
\newcommand{\camrev}[1]{#1}
\newcommand{\corrrev}[1]{#1}
\usepackage[disable]{todonotes}
\newcommand{\note}[4][]{\todo[author=#2,color=#3,size=\scriptsize,fancyline,caption={},#1]{#4}} %
\newcommand{\tiago}[2][]{\note[#1]{\textbf{Tiago}}{cyan!30}{#2}}
\newcommand{\clara}[2][]{\note[#1]{\textbf{Clara}}{orange!30}{#2}}
\newcommand{\Clara}[2][]{\clara[#1,inline]{#2}}

\newcommand{\response}[2]{\vspace{3pt}\hrule\vspace{3pt}\textbf{#1: }{#2}}
\newcommand{\defn}[1]{\textbf{#1}}
\newcommand{\defeq}{\mathrel{\stackrel{\textnormal{\tiny def}}{=}}}

\newcommand{\done}{{\color{red} (DONE)}\xspace}

\newcommand{\fasttext}{\myemph{fastText}\xspace}
\newcommand{\unilm}{\myemph{UnigramLM}\xspace}
\newcommand{\glotlid}{\myemph{GlotLID-M}\xspace}
\newcommand{\unilid}{\myemph{UniLID}\xspace}
\newcommand{\cld}{\myemph{CLD3}\xspace}

\newcommand{\likelihood}{\mathcal{L}}
\newcommand{\token}{v}
\newcommand{\stringunit}{s}
\newcommand{\str}{\mathbf{\stringunit}}
\newcommand{\tokens}{\mathbf{\token}}
\newcommand{\stralphabet}{\Sigma}

\newcommand{\toklength}{M}
\newcommand{\tokindex}{m}
\newcommand{\vocab}{\mathcal{V}}
\newcommand{\corpus}{\mathcal{C}}
\newcommand{\lang}{\ell}
\newcommand{\languages}{\Lambda}

\newcommand{\model}{p_{\theta}}
\newcommand{\detokfunc}{\ensuremath{\rotatebox[origin=c]{180}{$\tokfunc$}}}
\newcommand{\tokfunc}{\tau}

\DeclareMathOperator*{\argmax}{\mathrm{argmax}}
\newcommand{\allsegmentations}{\mathcal{T}}
\newcommand{\maxtoklength}{T_{\mathrm{max}}}

\newcommand{\smalldots}{...}

\newcommand{\unigramdist}{\boldsymbol{\phi}}
\newcommand{\unigramdistcur}{\unigramdist^{(n)}}
\newcommand{\tokenprob}{\unigramdist[\token]}

\newcommand{\countv}[2]{c_{#1}(#2)}          

\newcommand{\hatc}[2]{\widehat{c}_{#1}(#2)}

\icmltitlerunning{What Language is This? Ask Your Tokenizer.}

\begin{document}

\twocolumn[
  \icmltitle{What Language is This? Ask Your Tokenizer.}
  \icmlsetsymbol{equal}{*}

  \begin{icmlauthorlist}
    \icmlauthor{Clara Meister}{epfl}
    \icmlauthor{Ahmetcan Yavuz}{eth}
    \icmlauthor{Pietro Lesci}{camb}
    \icmlauthor{Tiago Pimentel}{eth}
  \end{icmlauthorlist}

  \icmlaffiliation{epfl}{EPFL}
  \icmlaffiliation{camb}{University of Cambridge}
  \icmlaffiliation{eth}{ETH Z\"urich}

  \icmlcorrespondingauthor{Clara Meister}{clara.meister@epfl.ch}
  \icmlkeywords{Language Identification, Tokenization, Language Modeling}

  \vskip 0.3in
]

\printAffiliationsAndNotice{}  %
\begin{abstract}
Language Identification (LID) is an important component of many multilingual natural language processing pipelines, where it facilitates corpus curation, training data analysis, and cross-lingual evaluation of large language models. Despite near-perfect performance on high-resource languages, existing systems remain brittle in low-resource and closely related language settings.
We introduce \unilid, a simple and efficient LID method based on the \unilm tokenization algorithm. 
In short, to predict a string's language label, we simply ask: under which language's unigram distribution is this string most likely? 
Our formulation is data- and compute-efficient,  supports incremental addition of new languages without retraining existing models, and can naturally be integrated into existing language model tokenization pipelines. Empirical evaluations against widely used baselines, including \fasttext, \glotlid, and \cld, show that \unilid achieves competitive performance on standard benchmarks, reaches 69\% accuracy with five labeled samples per language and 89\% with 25, and delivers large gains on fine-grained dialect identification.\footnote{Implementation at \href{https://github.com/Ahmetcanyvz/UNILID}{github.com/Ahmetcanyvz/UNILID}.}\looseness=-1
\end{abstract}

\section{Introduction}

Language Identification (LID) systems are an important component of today's natural language processing (NLP) pipelines.
They are particularly important for the training and evaluation of large multilingual language models, where they are used to: (i) construct labeled corpora with adequate coverage of the world’s languages, (ii) analyze the composition of large-scale training data,  and (iii) evaluate models' cross-lingual performance. 
As a result, reliable and efficient LID systems have become a vital tool for multilingual language modeling pipelines.

Simple and efficient algorithms have worked well for LID \cite{cavnar01,joulin-etal-2017-bag,lui-baldwin-2012-langid}; many regard it as a \enquote{solved} task since performance on high- and medium-resource languages is often near perfect \cite{mcnamee_solved,lid_survey}. 
However, LID systems often still perform poorly on low-resource languages, closely related language pairs, or fine-grained dialectal distinctions \citep[][\emph{inter alia}]{caswell-etal-2020-language,chifu-etal-2024-vardial,goot-2025-identifying}. 
Even the language identification systems in widest use fail to consistently identify less common languages \cite{kargaran-etal-2023-glotlid}. 
Large language models fare no better: when prompted to identify the language of a text, ChatGPT reaches near-zero F1 on more than half of the 670 languages evaluated by \citet{chen-etal-2024-fumbling}. 
Collectively, these findings suggest that LID remains far from solved.

In this work, we propose a simple and computationally efficient approach to LID that combines the generative modeling framework underlying the \unilm tokenization algorithm with the classical Bayes decision rule. 
Concretely, \unilm assumes that a string is generated as a sequence of latent subword units drawn independently from a unigram distribution. Rather than learning a single global distribution, we estimate a separate unigram distribution per language. 
Concretely, for each language, we find the most probable segmentation of the input string under that language's unigram distribution. The probability of that segmentation is our estimate of the string's likelihood under that language. Segmentation is therefore a language-dependent latent variable.
Finally, we apply Bayes’ rule to the resulting language-conditional likelihoods to obtain a posterior distribution over languages, and we assign the language label with the highest posterior probability.

The proposed method offers several advantages over prior LID approaches. 
While related to earlier generative LID models \cite{cavnar01,dunning1994statistical}, our method departs in a key respect by treating segmentation as a language-specific phenomenon rather than enforcing a fixed tokenization across languages.\clara{Maybe a bridge sentence like "this seems to be a key inductive bias"} 
The per-language distribution formulation further enables incremental extension to new languages or dialects without retraining existing models, and empirically, only a few samples per language are needed to learn language-specific parameters that achieve strong performance.
Finally, inference is computationally efficient: using clever dynamic programming tricks, inference can be done in an amount of time comparable to the inference step of the classic \unilm tokenization algorithm. %

Across a diverse set of benchmarks, \unilid achieves performance competitive with widely-used baselines while requiring substantially fewer labeled examples. The method is particularly effective in regimes where LID remains challenging. 
In our experiments subsampling training sets to mimic the low-resource setting, \unilid attains 69\% accuracy with five samples per language, 89\% at 25 samples and 93\% at 50 (\cref{tab:samples-accuracy}). 
On fine-grained dialect identification, \unilid improves macro F1 from 0.53 to 0.72 relative to a \fasttext baseline. 
Two simple calibration steps (\cref{sec:calibration}) raise macro F1 on the 1{,}940-language GlotLID-C test set significantly: On the full scored pool \unilid scores 0.933 and calibrated \unilid 0.956 (\cref{tab:lid_main}), again surpassing the \fasttext baseline. 
Taken together, these results suggest that treating segmentation as a language-specific phenomenon provides an effective foundation for language identification.

\newcommand{\lidfunc}{f_\mathtt{lid}}

\section{Language Identification}

Let $\str=\langle\stringunit_1, \dots, \stringunit_T\rangle$ be a string, \ie, a sequence of characters\footnote{The term \enquote{characters} denotes the atomic symbols of the input alphabet, encompassing both Unicode code points and bytes.}
from an alphabet $\stralphabet$, and $\languages$ a set of language labels. 
Closed-set LID is the task of assigning a language $\lang\in\languages$ to $\str$. 
This problem is often framed probabilistically, \ie,
\begin{equation}\label{eq:lang-id}
    \lidfunc(\str) = \argmax_{\lang\in\languages} \,\,p(\lang\mid\str)
\end{equation}
and the task then becomes approximating $p$ with a learned model $\model$. 
Two modeling paradigms are typically employed for solving this task. Discriminative approaches model this distribution using a score $f_\theta(\str,\lang)\in\R$, which is normalized using a softmax function:
\begin{equation}\label{eq:discriminative}
    \model(\lang \mid \str) = \frac{\exp\{f_\theta(\str,\lang)\}}{\sum_{\lang'\in\languages}\exp\{f_\theta(\str,\lang')\}}
\end{equation}
Generative approaches model this distribution via language-conditional likelihoods $\model(\str\mid \lang)$; the desired distribution can then be computed via the Bayes rule:\footnote{We assume a uniform prior over languages in $\languages$ and so drop the explicit $p(\lang)$ term to reduce clutter. }
\begin{equation}\label{eq:bayes}
\model(\lang \mid \str) = \frac{\model(\str \mid \lang)}{\sum_{\lang'\in\languages} \model(\str \mid \lang')}
\end{equation} 
Prior approaches to LID proposed different methods for parameterizing these distributions, as we discuss next.

\subsection{Prior Approaches}

\paragraph{\ngram Models.} 
Early work on LID established the effectiveness of using plain character-level \ngram statistics with the generative modeling approach of \cref{eq:bayes} \cite{cavnar01,dunning1994statistical}. Explicitly, they estimated language-conditional probabilities with character-level \ngram language models---often augmented with standard backoff or smoothing techniques \citep[\eg, Kneser--Essen--Ney smoothing;][]{kneser}.
These models set a long-standing baseline for robust LID, particularly on short or noisy texts \cite{Vatanen10LREC,baldwin-lui-2010-language}.

\paragraph{Discriminative Frameworks.}
In supervised settings, discriminative frameworks often showed improvements over earlier generative approaches, particularly when sufficient training data was available and languages were well represented \cite{lid_survey}.
The standard approach represents input strings as feature vectors encoding character \ngram statistics, such as term frequency scaled by inverse document frequency (TF-IDF) weights.
Formally, letting $\gamma(\str) \in \R^d$ denote this representation, a linear discriminative model scores each language with $f_\theta(\str, \lang) = \theta_{\lang}^{\top} \gamma(\str)$, instantiating \cref{eq:discriminative}, with $\theta_\lang$ learned by minimizing empirical cross-entropy over labeled data.
A canonical example is \fasttext \cite{joulin-etal-2017-bag}, which averages character $n$-gram embeddings into a single hidden representation and passes it to a linear classifier.
Its architecture is simple, but it has become a standard reference point in LID research: \myemph{OpenLID} and \glotlid pair it with large-scale, curated training data \cite{burchell-etal-2023-open,kargaran-etal-2023-glotlid}, and \myemph{ConLID} \cite{foroutan2025conlid} adds supervised contrastive learning on the same architecture.

\paragraph{Neural Approaches.} 
In more recent years, neural approaches have been applied to LID. Most are based on character- and byte-level neural sequence models, which arguably provide a more flexible alternative to bag-of-\ngram representations such as \fasttext. Character-level CNN and bidirectional RNN baselines learn feature representations directly from raw text, eliminating the need for explicit feature engineering. %
These systems have shown competitive performance, including for streaming or short text LID \cite{zhang_cnn_2015,belinkov-glass-2016-character,kocmi-bojar-2017-lanidenn}.
Google's Compact Language Detector v3 (\cld) uses a shallow feed-forward network over normalized character \ngram frequency features. %
Multilingual Transformer encoders (\eg, mBERT, XLM-R) reach high accuracy with task-specific fine-tuning, but their computational overhead and memory footprint often make them impractical for high-throughput or resource-constrained settings.

\subsection{Open Problems}

\paragraph{Varietal and dialectal discrimination.} 
Closely related languages or dialects (\eg, Bosnian, Croatian, and Serbian; Hindi and Urdu; or Syrian and Lebanese Arabic) share most of their orthography and much of their subword usage. As a result, current LID systems make more errors on these distinctions than on unrelated languages. Further, there is typically insufficient labeled training data to be able to learn these more fine-grained distinctions \cite{gaman-etal-2020-report,chifu-etal-2024-vardial}.

\paragraph{Low-resource performance.} 
As with many NLP datasets, there exists much more data for a core set of high-resource languages than for the long tail of low-resource ones.
Further, the data that does exist for these low-resource languages is often noisy, \eg, contaminated with HTML code or snippets of text from high-resource languages \cite{kreutzer-etal-2022-quality}.
Parameter estimates for these languages are often poor and, consequently, so is LID performance \cite{caswell-etal-2020-language,kargaran-etal-2023-glotlid}. 
Reweighting, data augmentation, and contrastive objectives have shown promising improvements but have not yet eliminated performance gaps across different resource levels.\cite{ren-etal-2022-effective,burchell-etal-2023-open,foroutan2025conlid}

\paragraph{Domain shift and orthographic noise.} 
Models trained on curated text collections often generalize poorly to informal or domain-mismatched inputs such as social media or transliterated scripts \cite{goot-2025-identifying,ojo2025diversbenchevaluatinglanguageidentification}. Many systems are also not robust to perturbations in orthography (\eg, diacritic omission in languages like French or Vietnamese), as these substantially alter character \ngram statistics, leading to pronounced drops in both accuracy and calibration \cite{caswell-etal-2020-language}.\looseness=-1

\paragraph{This work.} \corrrev{UniLID targets varietal discrimination and low-resource performance.} As we show in (\cref{sec:results}), it substantially improves dialect discrimination on DSL-ML 2024 (macro F1 0.53 $\rightarrow$ 0.72) and yields large sample-efficiency gains in the low-resource regime \corrrev{(69\% accuracy with five samples per language when the WiLI training set is subsampled)}. On out-of-domain inputs it gives partial improvements\corrrev{; on orthographic noise \unilid and \fasttext are roughly tied at corruption rates up to 10\%, and \fasttext degrades less at higher rates} (\cref{sec:results}).

\section{UnigramLM}

Tokenization is the process of segmenting strings into sequences of subword units, \ie, character spans referred to as \emph{tokens} in this context. %
Formally, let $\stralphabet$ be the alphabet of characters $\stringunit$, and $\vocab$ the vocabulary of tokens $\token$. 
Then, tokenization is the mapping of sequences of characters $\str=\langle \stringunit_1,\stringunit_2,\dots\rangle \in\stralphabet^*$ to sequences of tokens 
$\tokens=\langle \token_1,\token_2,\dots\rangle\in\vocab^*$, which we denote as 
$\tokfunc\colon \stralphabet^* \rightarrow \vocab^*$. This mapping is typically injective and thus reversible: the original string can be obtained from its tokenized form through a deterministic detokenization function $\detokfunc\colon \vocab^* \rightarrow \stralphabet^*$, \ie, $\str=\detokfunc(\tokfunc(\str))$. As each token (typically) represents a character span, the definition of $\detokfunc$ is often simple, reducing to concatenating the characters within each token together, \ie: $\detokfunc(\langle\token_1, \token_2, \ldots\rangle) \defeq \token_1 \circ \token_2 \circ \cdots$. 
For example, $\detokfunc(\langle\tokenprint{he},\tokenprint{llo}\rangle) = \chspan{hello}$. Notably, the detokenization function is not necessarily 1-to-1: different token sequences may yield the same string. Reusing the example above, we observe that $\detokfunc(\langle\tokenprint{hell},\tokenprint{o}\rangle)$ still detokenizes to $\chspan{hello}$.
The non-injectivity of $\detokfunc$ is an aspect of tokenization that we will later take advantage of in our LID approach. 
Going forward, we refer to $\tokens$ as a \defn{segmentation} of $\str$ if $\detokfunc(\tokens) = \str$ and define $\allsegmentations_{\vocab}(\str) = \{\tokens \corrrev{\in \vocab^*} : \detokfunc(\tokens) = \str\}$ as the set of all valid segmentations of $\str$ under vocabulary $\vocab$.

Numerous tokenization algorithms exist. 
These algorithms define a mapping function $\tokfunc$ and the method for learning the parameters of this mapping function from data. 
For example, the Byte-Pair Encoding \citep[BPE;][]{sennrich-etal-2016-neural} algorithm defines a $\tokfunc$ parametrized by a list of merges, and a method to learn this list. 
In this work, we focus on the \unilm algorithm \cite{kudo-2018-subword}, which will form the basis of our LID approach.

\subsection{Generative Model}

In broad strokes, \unilm casts tokenization as the uncovering of a latent segmentation: it assumes strings are simply the surface form of a latent sequence of tokens $\tokens$, where each token is drawn independently from a categorical (unigram) distribution.

Let $\unigramdist \in \Delta^{|\vocab|-1}$ denote a probability distribution over a vocabulary $\vocab$, where $\tokenprob$ gives the probability of token $\token$. By construction, $\tokenprob \geq 0$ and $\sum_{\token\in\vocab}\tokenprob = 1$. 
Under the \unilm framework, a token sequence of fixed length $\toklength$ is generated by independently sampling each token from this distribution, and the likelihood of a token sequence $\tokens = \langle \token_1, \smalldots, \token_M\rangle$ is the joint probability of those tokens:\footnote{We would need either an \textsc{eos} token (absorbing state) or a length prior in order to make this a valid probability distribution over $\vocab^*$. This detail is ignored in the standard \unilm implementation, and so we disregard it here as well for faithfulness to the original algorithm.}\clara{@Tiago, would it be correct to say in this footnote also that: in the context of comparing per-language string probabilities, if we assume eos is the same across languages, it would just cancel out? \response{tiago} Hmm, you mean if you assume p(eos) is fixed and constant across languages? Yeah, than I think it's true that it cancels out. Not sure if this is a reasonable assumption though... }
\begin{equation}\label{eq:pZ_single}
    \underset{\text{for}\,\,\tokindex = 1, \ldots, \toklength }{\token_{\tokindex} \stackrel{\text{i.i.d.}}{\sim} \mathrm{Categorical}(\unigramdist)} \, \implies p_{\unigramdist}(\tokens) \defeq \prod_{\tokindex=1}^{\corrrev{\toklength}} \unigramdist[\token_{\tokindex}]
\end{equation}
A string $\str$ is then deterministically generated by detokenization. Consequently, the probability of $\str$ is obtained by marginalizing over all valid segmentations:
\begin{equation}
    p_{\unigramdist}(\str) = \sum_{\tokens\in\allsegmentations_{\vocab}(\str)}p_{\unigramdist}(\tokens) 
\end{equation}
which formalizes the data-generating process of a string $\str$ under \unilm's assumptions. 
The posterior distribution over segmentations then follows naturally: %
\begin{equation}\label{eq:posterior}
    p_{\unigramdist}(\tokens\mid\str) = \frac{p_{\unigramdist}(\tokens)}{p_{\unigramdist}(\str)} \quad \text{for $\tokens \in \allsegmentations_{\vocab}(\str)$ and $0$ otherwise}
\end{equation}

Crucially, both  $\unigramdist$ and $\vocab$ are unknown; we only observe the strings resulting from the generative process. The \unilm algorithm estimates these parameters from text data by combining the Expectation–Maximization (EM) algorithm with an iterative vocabulary pruning procedure.

\subsection{Learning Model Parameters}\label{sec:learning-params}

The aim of the procedure proposed by the \unilm algorithm is to find the values of $\unigramdist$ and $\vocab$ that maximize data likelihood. 
Under the \unilm assumptions about the generative process of strings, the \enquote{complete} data consists of $(\str,\tokens)$ pairs, \ie, strings and the sequence of tokens that produced them. 
Since we only observe strings, and not their underlying segmentations, we cannot directly optimize for \emph{complete}-data log-likelihood; we must instead optimize for the \emph{observed}-data log-likelihood, where the observed data is our text corpus $\corpus=\{\str_k\}_{k=1}^K$:\clara{add something about a prior?}
\begin{equation}\label{eq:corpus-likelihood}
    \likelihood(\corpus; \unigramdist) \defeq \sum_{k=1}^K\,\log\!\!\sum_{\tokens\in \allsegmentations_{\vocab}(\str_k)} p_{\unigramdist}(\tokens) 
\end{equation}
The quantity in \cref{eq:corpus-likelihood} is difficult to optimize directly due to the log-sum structure. The EM algorithm establishes a relationship between this quantity and the expected value of the complete-data log-likelihood, which allows us to solve an easier problem in its stead. 
The EM algorithm alternates an E-step and an M-step until the estimates $\unigramdistcur$ converge. The E-step computes the expected complete-data log-likelihood of the observed corpus $\corpus$ under the current estimate $\unigramdistcur$. The M-step maximizes that expectation over $\unigramdist$. In the context of \unilm:\footnote{$\unigramdist^{(0)}$ can be initialized as \eg, the uniform distribution. See \citet{land2025piecesdoesunigramtokenization} for a discussion of this design choice. }
\begin{enumerate}[nosep,leftmargin=*]
    \item \textbf{E-Step.} Let $\countv{\token}{\tokens} \defeq \sum_{\tokindex=1}^{|\tokens|}\mathbbm{1}\{\token_\tokindex = \token\}$, \ie, the number of times token $\token\in\vocab$ appears in a segmentation $\tokens$.
    Given current parameters $\unigramdistcur$ and fixed $\vocab$, the E-step computes expected counts in our corpus as
    \begin{equation}
       \hatc{\token}{\corpus;\unigramdistcur}
        \defeq \sum_{k=1}^K 
        \sum_{\tokens\in\allsegmentations_{\vocab}(\str_k)}
        p_{\unigramdistcur}(\tokens \mid \str_k)\; \countv{\token}{\tokens}
    \end{equation}
    This computation is performed using dynamic programming, specifically via the forward-backward algorithm.
    \item \textbf{M-Step.} The M-step then computes the new parameter estimates as simply the normalized expected counts of each token:
    \newcommand{\unigramdistnewtoken}{\unigramdist^{(n+1)}[\token]}
    \begin{equation}\label{eq:mstep}
        \unigramdistnewtoken
        \;=\;
        \frac{\hatc{\token}{\corpus;\unigramdistcur}}
        {\sum_{\token'\in{\vocab}}\hatc{\token'}{\corpus;\unigramdistcur}}.
    \end{equation}
\end{enumerate}

For a proof of\tiago{Is "proof" accurate here? Or would maybe "For a longer exposition on why ..." be better?} why this is a valid approximation of maximum likelihood estimation (MLE) for a fixed $\vocab$, we refer the reader to \corrrev{\citet{dempster1977maximum} for the general EM result and to} the exposition on \unilm in \citet{meister2025unigramlm}.

Intuitively, this procedure resembles maximum likelihood estimation for a categorical distribution, but with a key difference: we cannot compute observed (``hard'') counts of token occurrences, since the segmentation of each string into tokens is latent.  
We instead compute expected (``soft'') token counts under the current model by marginalizing over all possible segmentations. 
In this way, the EM procedure integrates out segmentation uncertainty when estimating $\unigramdist$, yielding parameter estimates that reflect the full distribution over tokenizations rather than a single, fixed segmentation.

\paragraph{Learning $\vocab$.} 
The EM procedure above operates over a fixed vocabulary $\vocab$. Indeed, the validity of the approximation relies on the fact that $\vocab$ is fixed (and known). 
In the case of our language identification algorithm, we can pre-specify it as, \eg, the vocabulary of a chosen language model. 
If one wishes to also learn the vocabulary, the \unilm algorithm proposes a heuristic approach. Initially, $\vocab$ is defined as an oversized set consisting of the most frequently occurring substrings in the corpus; this set can be found efficiently using Enhanced Suffix Array-based algorithms \citep[\eg,][]{Abouelhoda2002TheES}.
The vocabulary is then \enquote{learned} by iterating between: (i) holding it fixed and estimating $\unigramdistcur$ via EM; (ii) a pruning step, where the tokens whose removal lowers corpus log-likelihood the least, evaluated at the current estimate $\unigramdistcur$, are removed.
Via this pruning step, the algorithm ultimately trims the vocabulary down to the user's desired size.\looseness=-1

\subsection{Tokenizing Text}\label{sec:inference}

\newcommand{\unigramdistfinal}{\widehat{\unigramdist}}

At inference time (\ie, when turning text into tokens), the \unilm tokenization algorithm finds the most likely segmentation of $\str$ under the learned parameters $\unigramdistfinal = \unigramdist^{\corrrev{(R)}}$, \corrrev{the estimate after the final EM round $R$}, as:\footnote{In the original \unilm algorithm, the most-probable segmentation is chosen rather than the one that has the highest marginal probability. We compared both strategies at inference on GlotLID-C: marginalizing scored 0.002 higher macro F1 and the same accuracy to three decimals (no uncertainty estimate for this comparison) but took roughly 1.7 times as much time (see \cref{app:viterbi_vs_marginal}).}
\looseness=-1%
\begin{subequations}\label{eq:approx-inference}
\begin{align}
    \tokfunc_{\unigramdistfinal}(\str) 
    &\stackrel{\phantom{\text{(\cref{eq:posterior})}}}{=} \argmax_{\tokens \in \vocab^*} p_{\unigramdistfinal}(\tokens\mid\str) \\
    &\stackrel{\text{(\cref{eq:posterior})}}{=} \argmax_{\tokens \in \allsegmentations_{\vocab}(\str)} p_{\unigramdistfinal}(\tokens)\label{eq:drop_strprob}\\
    &\stackrel{\text{(\cref{eq:pZ_single})}}{=} \argmax_{\tokens \in \allsegmentations_{\vocab}(\str)} \prod_{\tokindex=1}^{|\tokens|}\unigramdistfinal[\token_\tokindex] \label{eq:target}
\end{align}
\end{subequations}
where we dropped $p_{\unigramdistfinal}(\str)$ in \cref{eq:drop_strprob} since it does not depend on $\tokens$.
The exact solution to \cref{eq:target} can be found efficiently using Viterbi decoding, which runs in $\mathcal{O}(T\cdot \maxtoklength)$ time, where $T$ is the length of the input string $\str$ and $\maxtoklength$ is the length of the longest token $\token\in\vocab$.

\section{\unilid}\label{sec:unilid}

\newcommand{\unigramdistlang}{\widehat{\unigramdist}_{\lang}}
\newcommand{\punigramlang}{p_{\unigramdistlang}}
\newcommand{\unigramdistnewlang}{\unigramdist^{(n+1)}_\lang}

We propose a generative modeling approach for the LID task using the \unilm framework.
In short, we define a language-conditional distribution $\model(\str\mid\lang)$ for each language $\lang$ by learning unigram distribution parameters per language, albeit with a shared vocabulary across languages. We then compute language label probabilities, as in  \cref{eq:bayes}, applying the Bayes' decision rule to ultimately assign the language label.
In standard terms, \unilid is a multinomial naive Bayes classifier over subword tokens in which the segmentation is latent rather than a fixed preprocessing step. 
The main novelty here is that the segmentation is estimated
separately for each language. 

\subsection{Base Algorithm}
The first step in \unilid consists of language-conditional distribution estimation. 
Let $\vocab$ be the vocabulary of a base tokenizer.\footnote{This does not need to be a \unilm tokenizer. Any tokenizer that uses a fixed vocabulary can be used.} 
We fix $\vocab$ as our shared-across-languages vocabulary, but assume there exists, for each language $\lang$, a distinct unigram distribution $\unigramdistlang$ over $\vocab$.
\clara{Is it confusing that we sometimes use 'per-language' and sometimes 'language-conditional'? \response{tiago} So far, I think it's ok. } 
Given a corpus $\corpus^\lang$
consisting of strings from language $\lang$, we then learn each of these distributions' parameters $\unigramdistlang$ using the EM procedure described in \cref{sec:learning-params}.\clara{it might be the case that for this problem, we're guaranteed to find the global optimum. Look into that... \response{tiago} Because it's a `simple' distribution? \response{clara}{I guess it would be because of the properties of the objective}}  In other terms, similarly to as in standard \unilm, we compute an approximate maximum likelihood estimate by maximizing the observed-data log-likelihood, albeit on only a subset of the data: 
\begin{equation}
    \unigramdistlang \corrrev{\approx} \argmax_{\unigramdist} \,\,\likelihood(\corpus^\lang; \unigramdist)
\end{equation}
where $\unigramdistlang$ is the estimate at which the EM iteration of \cref{sec:learning-params} converges, and the approximation is the one EM makes. 
As in the standard \unilm formulation, this procedure integrates over segmentation uncertainty\clara{is this mathematically accurate? \response{tiago} does this mean it marginalises over possible segmentations? I guess it's correct if yes, but a bit of a complicated way of saying that. } when estimating parameters. 
Because data from each language is processed separately, each language gets its own token frequency profile---and consequently its own set of optimal segmentations---despite sharing a common vocabulary across languages. We discuss several design choices we make in our practical implementation of \unilid (\eg, initialization of $\unigramdistlang$ for the EM loop and floor values for token probabilities) in \cref{sec:exp_setup}.

\paragraph{Inference.} At inference time, we use the inference procedure of \unilm discussed in \cref{sec:inference} to find the most probable segmentation of a string under each language's $\unigramdistlang$. 
Importantly, this means that the most probable segmentation $\tokfunc_{\unigramdistlang}(\str)$ can vary across languages. 
This yields a flexible model where the probability of a string reflects language-specific token frequencies, despite sharing a common vocabulary.  
We then approximate the probability of a string under language $\lang$ as the probability of its most probable segmentation in that language: $\model(\str\mid\lang) = p_{\unigramdistlang}(\tokfunc_{\unigramdistlang}(\str))$. Finally, we assign the language label that optimizes the posterior probability, \ie, \cref{eq:lang-id}. We write the instantiation of  \cref{eq:lang-id} here explicitly for the reader's convenience:%
\begin{align}
    \lidfunc(\str) = \argmax_{\lang\in\languages} \corrrev{\model}(\lang\mid \str)
   =\argmax_{\lang\in\languages}p_{\unigramdistlang}(\tokfunc_{\unigramdistlang}(\str)) \nonumber
\end{align}
The second equality follows from our assumption of a uniform prior over $\languages$ (\cref{eq:bayes}).

\subsection{Calibrated \unilid}
\label{sec:calibration}
\defn{Calibrated \unilid} is \unilid with the log-probability
assigned to unseen tokens capped at one shared constant and with predictions
reassigned when their log-probability margin (defined below) is small, both
applied at inference time. In machine learning the word calibration usually denotes
matching a system's predicted probabilities to observed frequencies; we use
calibrated \unilid as a name for the system just described, and we report no
such reliability measurement.
An error analysis of \unilid on GlotLID-C shows that misclassifications
concentrate on predictions \emph{into} two identifiable groups of languages:
languages with fewer than 18{,}000 training samples, whose estimates
$\unigramdistlang$ remain close to their initialization and therefore assign
moderate probability to strings from many languages, and a group of four
larger languages whose estimated distributions have unusually high entropy
for their script and which \camrev{receive} misclassified samples for the same reason
(the identification criteria are stated in \cref{app:protocol}).
On the scored GlotLID-C test pool (\cref{app:protocol}), 32.5\% of
\unilid's false positives (578{,}270 of 1{,}779{,}499) are predictions into
these two groups, whose 1{,}084 languages together hold 3.8\% of the test
lines by gold label. Both patterns
arise from under-fit distribution estimates. Calibrated \unilid
corrects for the two patterns at inference time.

The first correction concerns the tokens in $\vocab$ that never occur
in $\corpus^\lang$. Base \unilid assigns every such token the smallest value of
$\log\unigramdistlang$ over $\vocab$, and that value varies with the size of
$\corpus^\lang$: across the 1{,}940 languages it decreases by 2.04 nats for each
tenfold increase in the number of training tokens.\clara{Give intuition here}  Since \cref{eq:lang-id} compares scores across languages, this variation acts as a language-dependent offset on every unseen token. Calibrated \unilid lowers the value to one shared constant $c$ wherever it exceeds $c$, that is, $\log\unigramdistlang[\token] \defeq \min\bigl(\log\unigramdistlang[\token],\, c\bigr)$ for every $\token$ unseen in $\corpus^\lang$. 
The constant $c$ is unrelated to the expected count $\widehat{c}_{\token}$ of \cref{sec:learning-params}. For 1{,}655 of the 1{,}940
languages the value lies above $c$ and is set to $c$; for the remaining 285 it
already lies below $c$ and is unchanged.\footnote{The correction is applied without renormalizing
$\unigramdistlang$, so the corrected per-language values are unnormalized
scores; the decision rule compares them as in \cref{eq:bayes}, and the
posterior interpretation holds up to this approximation.}

The second correction re-examines predictions into the
two identified groups whose log-probability margin is small. Let \corrrev{$g_\lang(\str)$} $= \log
\punigramlang(\tokfunc_{\unigramdistlang}(\str))$ be the score assigned by
language $\lang$ and \corrrev{$\hat\lang = \argmax_\lang g_\lang(\str)$} the
prediction. The \defn{log-probability margin} of the prediction is the
top language's log-probability score minus the runner-up's, that is,
$g_{\hat\lang}(\str)$ minus the second-largest value of
$g_\lang(\str)$ over $\lang\in\languages$. If $\hat\lang$ belongs to either group and its log-probability margin is smaller than a per-language threshold $\delta_{\hat\lang}$, the prediction is reassigned to the highest-ranked of the top 5 languages that are \emph{not} in either group.\footnote{We also set an absolute threshold: the score under the replacement label must be within a fixed number of natural-log units of that of the top label.} If no candidate qualifies, the prediction is
unchanged. Calibrated \unilid is therefore a two-stage rule: it takes the argmax of the corrected scores, and then reassigns that prediction under the conditions just stated.

\paragraph{Setting hyperparameters.} The threshold $\delta_\lang$ is set from language $\lang$'s own
training data. 
We take the training examples on which $\lang$ is the
top-scoring language, compute the log-probability margin of each, and set $\delta_\lang$ to a low percentile of those margins. 
Setting $\delta_\lang$ at the $q$th percentile means that $q$ percent of those training lines fall below it and would be re-examined.\footnote{One threshold shared by all languages actually leads to a higher aggregate score, but hurts the lower-resource language more compared to a per-language threshold.} 
The remaining constants can be chosen using a validation set, as we do in our experiments (see \cref{app:protocol}).  Apart from the high-entropy-group identification, both corrections depend only on shared constants and on each
language's own data, preserving \unilid's ability to add languages incrementally.

\paragraph{Computational Complexity.}
At training time, \unilid estimates a separate unigram distribution for each language by running the EM procedure over the corresponding language-specific corpus. 
For a corpus $\corpus^\lang$ of total character length \camrev{$T_\lang$} (\eg, number of Unicode code points or bytes), each EM iteration requires computing expected token counts via the forward--backward algorithm, which runs in \camrev{$\mathcal{O}(T_\lang \cdot \maxtoklength)$}, where again $\maxtoklength$ denotes the maximum token length. Parameter updates are linear in the vocabulary size $|\vocab|$. 
Since per-language distribution parameters are estimated independently, training parallelizes trivially across languages. At inference time, \unilid uses the same Viterbi decoding as \unilm:
first building a lattice that enumerates all valid segmentations of a string $\allsegmentations_\vocab(\str)$ and then finding the maximum probability option among them. 
The lattice needs only to be constructed once per string, regardless of the number of language-conditional distributions. 
Finding the optimal segmentation of a string and its probability for each language then takes $\mathcal{O}(|\languages| \cdot E)$ time, where $E$ is the number of edges in the lattice.
Finally, we need a $\mathcal{O}(|\languages|)$ pass to evaluate the Bayes decision rule.
The number of edges $E$ in a lattice is \camrev{$T\cdot b$}, where $b$ is the lattice's branching factor, \ie, how many tokens in the vocabulary start at a given string position and are part of a valid segmentation (typically 1-5), so inference takes \camrev{$\mathcal{O}(|\languages|\cdot T\cdot b + T \cdot\maxtoklength)$}.
We compare this cost with \fasttext and \cld in \cref{app:efficiency} and report empirical latency in \cref{sec:results}.

\section{Experimental Setup}\label{sec:exp_setup}

We evaluate the performance of \unilid across several prominent LID benchmarks.
We note that the majority of our training protocol and evaluation paradigms follow prior work \cite{kargaran-etal-2023-glotlid,foroutan2025conlid}. 
Our experiments compare against widely-used baselines: \fasttext, \cld, and \glotlid. We perform analyses to show data efficiency, robustness to input length, and the impact of using different tokenizer vocabularies.\looseness=-1

\subsection{Datasets}

We choose benchmarks that cover a spectrum of LID regimes, ranging from broad linguistic coverage and low-resource settings to fine-grained dialect identification: GlotLID-C \citep{kargaran-etal-2023-glotlid} for large-scale coverage of
the long tail of nearly 2{,}000 language--script labels;
UDHR \citep{Vatanen10LREC} and FLORES-200 \citep{nllb2022}
for content-controlled cross-lingual evaluation via parallel text across
hundreds of languages;
DSL-ML 2024 \citep{chifu-etal-2024-vardial} for fine-grained dialect
identification across 14 regional varieties with high lexical overlap;
Tatoeba \citep{tiedemann-2020-tatoeba} for robustness on short, noisy,
community-contributed text;
and WiLI-2018 \citep{Thoma2018TheWB} for controlled ablations, as its
balanced 500/500 train/test split across 235 languages supports
controlled comparisons across different dataset stratifications  at
manageable scale. We use official train/test splits where available; full
per-dataset descriptions, including sample counts and split construction,
are deferred to \cref{app:datasets}.
\subsection{Baselines}

We compare \unilid against the following systems. For a fair comparison, we hold the training dataset constant across systems that we train ourselves. 
For systems that did not release reproduction code, or for which we could not reproduce the associated paper's reported performance, we use the released checkpoints. 

\paragraph{\fasttext.} 
We train supervised \fasttext  \cite{joulin-etal-2017-bag} classifiers using the author's Python package\footnote{\href{https://pypi.org/project/fasttext}{\myemph{https://pypi.org/project/fasttext}}.} \corrrev{with the package's default configuration, apart from a sweep over selected hyperparameters to improve performance (\cref{app:fasttext_epochs})}. Models were trained for 100 epochs.\footnote{While the authors' base setup uses fewer epochs, models trained for fewer epochs seriously underperformed on benchmarks in our experiments. We tuned \fasttext extensively, and 100 epochs gave the best performance in every setting; see \cref{app:fasttext_epochs} for results on our \fasttext hyperparameter sweep on DSL-ML 2024.}

\paragraph{\cld (pre-trained).} Google's Compact Language Detector v3 (\cld) is a lightweight neural network that supports 107 languages and is optimized for short text segments and low-latency inference. 
We use the publicly available pre-trained model from their Python package.\footnote{\href{https://github.com/google/cld3}{\myemph{https://github.com/google/cld3}}.}
Its language coverage does not fully overlap with that of our test sets. 
Thus, for a fairer comparison, we provide systems' results computed over only the subset of languages covered by \cld (right side of \cref{tab:lid_main}). In those columns \unilid and \fasttext are each fitted to the subset's languages. \Cref{tab:cld3_refit} reports what the full 1{,}940-language \unilid and its calibrated form score on the same slices.

\paragraph{\glotlid (pre-trained).}
\glotlid is the model provided with the GlotLID-C benchmark dataset. Architecturally, it utilizes the \fasttext framework but differentiates itself through its training data curation.  We use their openly available checkpoint, hosted on the same platform as the dataset. 
The exact training splits for \glotlid were not released but the
authors state that UDHR and FLORES-200 are not part of the training
set,\footnote{Their contamination check finds that for 57 languages,
$\leq 10\%$ of UDHR sentences appear in training; see \S3 of
\citet{kargaran-etal-2023-glotlid}. Notably though, for those 57 languages, the UDHR column of
\cref{tab:lid_main} favors \glotlid.} so we restrict GlotLID-M evaluations
to these two benchmarks. 

\begin{table*}[ht]
\centering
\small
\caption{Classification results on benchmark evaluation sets.
\unilid variants and \fasttext are trained using the GlotLID-C corpus.
The left columns show results on the full test sets of each benchmark for models that have full coverage of those benchmark's language label set. The right columns show results on the subsets of the benchmarks that \cld has label coverage for. A dash marks a system without coverage of a column's label set; bold marks the best value in a column. \Cref{tab:cld3_refit} reports what the full 1{,}940-language \unilid (i.e., without subsetting to the benchmark's label set) and its calibrated form score on the same benchmarks.}
\label{tab:lid_main}
\setlength{\tabcolsep}{5pt}
\renewcommand{\arraystretch}{1.15}
\adjustbox{max width=0.85\textwidth}{
\begin{tabular}{l cc cc cc | cc cc cc}
& \multicolumn{6}{c}{\normalsize \textbf{Full GlotLID-C label set}} &
 \multicolumn{6}{c}{\normalsize \textbf{\cld label set}}\\
\toprule
\textbf{Method} &
\multicolumn{2}{c}{\begin{tabular}[c]{@{}c@{}}\textbf{GlotLID-C test}\\ \small(1940 labels)\end{tabular}} &
\multicolumn{2}{c}{\begin{tabular}[c]{@{}c@{}}\textbf{UDHR}\\ \small(366 labels)\end{tabular}} &
\multicolumn{2}{c}{\begin{tabular}[c]{@{}c@{}}\textbf{FLORES-200}\\ \small(190 labels)\end{tabular}} &
\multicolumn{2}{c}{\begin{tabular}[c]{@{}c@{}}\textbf{GlotLID-C test}\\ \small(83 languages)\end{tabular}} &
\multicolumn{2}{c}{\begin{tabular}[c]{@{}c@{}}\textbf{UDHR}\\ \small(80 languages)\end{tabular}} &
\multicolumn{2}{c}{\begin{tabular}[c]{@{}c@{}}\textbf{FLORES-200}\\ \small(77 languages)\end{tabular}} \\

& F1$\uparrow$ & FPR$\downarrow$
& F1$\uparrow$ & FPR$\downarrow$
& F1$\uparrow$ & FPR$\downarrow$
& F1$\uparrow$ & FPR$\downarrow$
& F1$\uparrow$ & FPR$\downarrow$
& F1$\uparrow$ & FPR$\downarrow$ \\
\midrule

\cld 
& -- & -- 
& -- & -- 
& -- & -- 
& .906 & 5.48e-4
& .965 & 2.52e-4
& .978 & 2.22e-4 \\

\glotlid
& -- & --
& \textbf{.871} & \textbf{1.37e-4}
& \textbf{.968} & \textbf{1.35e-4}
& -- & --
& .994 & \corrrev{\textbf{2.09e-5}}
& \textbf{.999} & \textbf{6.25e-6} \\

\fasttext
& \camrev{.944} & 2.71e-5
& .855 & 1.87e-4
& .938 & 2.38e-4
& \textbf{.990} & \textbf{7.92e-5}
& \textbf{.998} & 2.98e-5
& .998 & 2.96e-5 \\

\midrule
\unilid
& \corrrev{.933} & \corrrev{2.02e-5}
& \corrrev{.856} & \corrrev{1.52e-4}
& \corrrev{.931} & \corrrev{2.83e-4}
& \corrrev{.974} & \corrrev{1.52e-4}
& \corrrev{.995} & \corrrev{5.73e-5}
& \corrrev{.997} & \corrrev{3.93e-5} \\

\unilid (calibrated, \cref{sec:calibration})
& \corrrev{\textbf{.956}} & \camrev{\textbf{1.77e-5}}
& \corrrev{.842} & \corrrev{2.03e-4}
& \corrrev{.932} & 2.91e-4
& \corrrev{.981} & \corrrev{1.48e-4}
& \corrrev{.993} & \corrrev{7.57e-5}
& \corrrev{.997} & \corrrev{3.29e-5} \\

\unilid-Mistral-Nemo
& .912 & \camrev{1.84e-5}
& .836 & 1.79e-4
& .923 & 3.22e-4
& .972 & 1.53e-4
& .992 & \camrev{1.03e-4}
& .994 & 8.05e-5 \\

\unilid-DeepSeek3.2
& .909 & 2.08e-5
& .836 & 1.78e-4
& .914 & 3.42e-4
& .971 & 1.66e-4
& .990 & 1.33e-4
& .994 & 7.92e-5 \\

\unilid-Qwen3
& .904 & 2.55e-5
& .829 & 1.91e-4
& .904 & 3.80e-4
& .964 & 2.27e-4
& .984 & 2.11e-4
& .989 & 1.46e-4 \\

\bottomrule
\end{tabular}}
\end{table*}

\subsection{Evaluation Metrics}

We report standard evaluation metrics for language identification, namely the macro-averaged F1 score and the macro-averaged false positive rate (FPR)\camrev{, defined as}
{\color{red}\begin{equation}
    \mathrm{F1} = \frac{1}{|\languages|} \sum_{\lang \in \languages} \frac{2 \cdot \mathrm{P}_{\lang} \cdot \mathrm{R}_{\lang}}{\mathrm{P}_{\lang} + \mathrm{R}_{\lang}},\,\, \mathrm{FPR} = \frac{1}{|\languages|} \sum_{\lang \in \languages} \frac{\mathrm{FP}_{\lang}}{\mathrm{FP}_{\lang} + \mathrm{TN}_{\lang}}
\end{equation}}%
where $\mathrm{P}_{\lang}$ and $\mathrm{R}_{\lang}$ denote the precision and recall for language $\lang$, $\mathrm{FP}_{\lang}$ counts the test samples of other languages predicted as $\lang$, and $\mathrm{TN}_{\lang}$ counts the test samples of other languages not predicted as $\lang$. Macro-averaging ensures that each language or dialect contributes equally to the overall score, regardless of class frequency. For analyses of data efficiency and robustness on the WiLI dataset, we additionally report classification accuracy.

\subsection{\unilid Training}
For each benchmark, we learn language-conditional unigram distributions using the EM procedure of the \unilm tokenization algorithm, described in \cref{sec:learning-params}.  
In all experiments, we initialize the unigram distribution as the uniform distribution and run EM for $20$ rounds.\footnote{We observed empirically that unigram distributions always converged within 20 rounds (\ie, the total variation distance between subsequent rounds fell below $10^{-6}$.}
Unseen tokens are assigned a corpus-size-dependent minimum, as implemented in the SentencePiece library. 
For calibrated \unilid (\cref{sec:calibration}), we use the shared unseen-token constant $c = -17$ (natural log). 
We set each log-probability margin threshold $\delta_\lang$ at the $q_\lang = 5\,(1 - N_\lang/18{,}000)$ percentile of language $\lang$'s own training margins for the languages below 18{,}000 samples, and at a fixed 5th percentile for the four high-entropy languages of \cref{sec:calibration}.
The full protocol, remaining constants, and their provenance are in \cref{app:protocol}.
We consider two variants of \unilid: one that uses a dedicated base tokenizer trained on the LID training set, and one that uses the vocabulary of a pretrained language model; more details below.

\paragraph{\unilid.}
The base version of \unilid trains a tokenizer on the training split of each benchmark dataset; for GlotLID-C, the base tokenizer is trained on a subsample of 1{,}000 samples per language (\cref{tab:training_time}). We train a byte-level tokenizer using the standard \unilm tokenization algorithm. Our implementation closely follows the design choices of the SentencePiece library \cite{kudo-richardson-2018-sentencepiece}, which provides the original implementation of \unilm. Unless otherwise specified, we use a vocabulary size of $100\text{k}$; ablation experiments indicate that performance is relatively insensitive to this choice (see \cref{sec:ablations}).\clara{potentially add details of pretokenization choices: ByteLevel(add\_ prefix\_space=True, use\_regex=True)}

\paragraph{\unilid-X.}
This variant uses the vocabulary of a pretrained base tokenizer and learns only the per-language unigram distributions. 
This setting evaluates \unilid’s ability to operate with pretrained, task-agnostic vocabularies. 
We report results using the \myemph{Mistral-Nemo-Base-2407}, DeepSeek3.2 and Qwen3 tokenizers in \cref{tab:lid_main}, and \cref{tab:unilid_llm_comparison} reports six pretrained base tokenizers on WiLI.

\begin{table*}[t]
\centering
\small
\caption{Per-dialect and macro-averaged F1 scores on the DSL-ML 2024 task for French (FR), Spanish (ES), Portuguese (PT), English (EN), and the Bosnian/Croatian/Montenegrin/Serbian family.}
\label{tab:per_language_f1}
\setlength{\tabcolsep}{4pt}
\renewcommand{\arraystretch}{1.15}
\adjustbox{max width=0.7\textwidth}{
\begin{tabular}{l *{4}{c} | *{2}{c} | *{2}{c} | *{2}{c} | *{4}{c} | c}
\toprule
& \multicolumn{4}{c}{\textbf{French}} & \multicolumn{2}{c}{\textbf{Spanish}} & \multicolumn{2}{c}{\textbf{Portuguese}} & \multicolumn{2}{c}{\textbf{English}} & \multicolumn{4}{c}{\textbf{BCMS}} & \\
\cmidrule(lr){2-5} \cmidrule(lr){6-7} \cmidrule(lr){8-9} \cmidrule(lr){10-11} \cmidrule(lr){12-15}
\textbf{Model} & BE & FR & CA & CH & ES & AR & BR & PT & US & GB & SR & HR & BS & ME & \textbf{Macro} \\
\midrule
\unilid & .595 & .288 & .798 & \textbf{.456} & \textbf{.867} & \textbf{.832} & \textbf{.816} & \textbf{.724} & \textbf{.855} & \textbf{.815} & \textbf{.929} & \textbf{.788} & \textbf{.609} & \textbf{.750} & \textbf{.723} \\
\fasttext & \textbf{.697} & \textbf{.434} & \textbf{.811} & .444 & .833 & .686 & .729 & .609 & .815 & .561 & .833 & .000 & .000 & .000 & .532 \\
\bottomrule
\end{tabular}%
}
\end{table*}

\begin{table}[t]
\centering
\small
\caption{Macro-averaged F1 and false positive rate (FPR) of \fasttext and \unilid trained on WiLI and evaluated on the Tatoeba and UDHR benchmarks.}
\label{tab:tatoeba_udhr_comparison}
\setlength{\tabcolsep}{7pt}
\adjustbox{max width=\linewidth}{
    \begin{tabular}{lcccc}
    \toprule
    \textbf{Method} &
    \multicolumn{2}{c}{\textbf{Tatoeba} (201 langs)} &
    \multicolumn{2}{c}{\textbf{UDHR} (142 langs)} \\
    & \textbf{Macro F1}$\uparrow$ & \textbf{Macro FPR}$\downarrow$
    & \textbf{Macro F1}$\uparrow$ & \textbf{Macro FPR}$\downarrow$ \\
    \midrule
    \fasttext & 0.160 & 3.58e-3 & 0.849 & 6.07e-4 \\
    \unilid & \textbf{\corrrev{0.420}} & \textbf{\corrrev{9.23e-4}} & \textbf{\corrrev{0.866}} & \textbf{\corrrev{5.86e-4}} \\
    \bottomrule
    \end{tabular}%
}
\end{table}
 
\section{Results}\label{sec:results}

\paragraph{Base Results.} 
We first evaluate \unilid on standard LID benchmarks: GlotLID-C, UDHR, and FLORES-200. \unilid is not always the top-performing system in terms of F1. 
Notably though, on the largest evaluation set (full GlotLID-C test), \unilid reduces FPR by roughly 25\% compared to \fasttext (2.02e-5 vs 2.71e-5), a property that can be more critical than F1 in many LID applications. For example, in web-scale crawling, a poor FPR for a low-resource language can lead to \corrrev{a training corpus in which misclassified lines outnumber correct ones} (common languages misclassified as rare ones).
In subsequent sections, we compare primarily against \fasttext. It is the only of these systems for which we can control training data (\glotlid is an instantiation of \fasttext, just with a particular dataset); we cannot control for test-set contamination or sample efficiency with the other systems.

\paragraph{Effect of Calibrated \unilid.}
On the part of our GlotLID-C test pool that was not used in any constant selection for calibrated \unilid, calibrated \unilid (\cref{sec:calibration}) raises macro
F1 from 0.916 to 0.949 , against \fasttext's
0.933 on that same set (\cref{tab:calibrated_heldout}). See \cref{app:protocol} for full details. Two of the calibration constants were selected on data inside the full scored pool, so we take this held-out figure as the fair assessment, devoid of any data leakage. 
On the full scored pool, \unilid scores 0.933 macro F1 and
calibrated \unilid 0.956, above every system in \cref{tab:lid_main} with full
GlotLID-C label coverage; on that pool, calibrated \unilid's macro FPR is 1.77e-5 compared to \unilid's 2.02e-5. 
The gains provided by calibrated \unilid are actually in the regimes where
\cref{tab:calibrated_views} reports \unilid's \corrrev{lowest F1}:  mean per-language F1 rises from \corrrev{0.596} to \corrrev{0.781} for languages with fewer than 500
training samples, and from \corrrev{0.676} to \corrrev{0.893} for languages with 500 to 1{,}000
training samples (scored GlotLID-C test pool). 
On FLORES-200, whose 190 labels are mostly higher-resource languages, macro F1 changes
from only 0.931 to 0.932. For this benchmark in particular, the re-examination component of calibrated \unilid changes few predictions: it re-examines 1{,}337 of 192{,}280 lines and reassigns 749.
\Cref{app:protocol} reports the same comparison on a
balanced draw of GlotLID-C.
On out-of-domain web-domain data (CommonLID, \cref{tab:commonlid}),
\corrrev{calibrated \unilid} raises macrolanguage-aware accuracy from \corrrev{0.848} to \corrrev{0.862}.

\paragraph{Dialect Differentiation.}
\cref{tab:per_language_f1} reports results on the DSL-ML 2024 benchmark. 
We compare \unilid against the organizers' SVM baseline on the same dev
split, scored with the shared task's official scoring script
\cite{chifu-etal-2024-vardial}. We make one change to ensure a protocol-matched comparison. 
That script averages F1 under a multi-label protocol on a single language group, while \unilid is scored single-label, so we change to scoring the baseline single-label instead.\footnote{Still, \unilid predicts over all 14 labels while the baseline chooses
among the 2 or 4 labels of a single group, so confusion across groups is by design not possible for the baseline.} 
Against the baseline, \unilid achieves better F1  on English
(0.835 vs. 0.775), Spanish (0.850 vs. 0.726) and Portuguese (0.770 vs. 0.667), and lower on French (0.534 vs. 0.638) and BCMS (0.769 vs. 0.826). 
We do not compare
against the shared task's ranked submissions, which are scored on a private
test split under the multi-label protocol. 
It also substantially outperforms \fasttext on macro F1, improving from 0.53 to 0.72. Part of that difference is the 0.00 F1 \fasttext obtains on HR, BS, and ME. This poor performance may in part be due to the small amount of training data for those languages (see \cref{app:datasets} for the dataset breakdown), a setting in which \fasttext also scores low (\cref{fig:samples-accuracy}: 10.5\% at five samples per language). 
The three labels on which the two systems differ most also have the fewest dev instances: 16 for HR, 16 for BS and 4 for ME (\cref{tab:dialect_stats}). 
We hypothesize that \unilid's strength in these settings is partially because its vocabulary $\vocab$ provides a structural inductive bias, so only per-language token frequencies must be estimated from the data, whereas \fasttext must adjust high-dimensional embeddings for millions of $n$-gram spans. Generative classifiers are also known to be more sample-efficient than discriminative ones when training data is scarce \cite{ng2001discriminative}, and \unilid is generative while \fasttext is discriminative. Neither explanation is isolated by an experiment here.

\begin{figure}[t]
    \centering
    \adjustbox{max width=0.8\linewidth}{
    \begin{tikzpicture}
    \begin{axis}[
        width=\linewidth,
        height=0.45\linewidth,
        grid=both,
        symbolic x coords={5,10,25,50,100,200,300,400,500},
        xtick={5,10,25,50,100,200,300,400,500}, 
        enlarge x limits=0.05,
        ymin=0, ymax=100,
        xlabel={Samples per Language},
        ylabel={Accuracy (\%)},
        legend pos=south east,
        legend cell align=left,
        tick align=outside,
    ]

    \addplot+[
        mark=*,
        line width=1pt,
        error bars/.cd,
            y dir=both,
            y explicit,
            error bar style={line width=1.2pt},
            error mark options={
                rotate=90,
                mark size=4pt,
                line width=1.2pt
            },
    ]
    coordinates {
        (5,10.53) +- (0,0.43)
        (10,0.85) +- (0,0.00)
        (25,13.25) +- (0,0.53)
        (50,67.79) +- (0,0.52)
        (100,85.70) +- (0,0.40)
        (200,91.70) +- (0,0.04)
        (300,93.20) +- (0,0.04)
        (400,94.03) +- (0,0.03)
        (500,94.55) +- (0,0.06)
    };
    \addlegendentry{\fasttext}

    \addplot+[
        mark=square*,
        line width=1pt,
        error bars/.cd,
            y dir=both,
            y explicit,
            error bar style={line width=1.2pt},
            error mark options={
                rotate=90,
                mark size=4pt,
                line width=1.2pt
            },
    ]
    coordinates {
        (5,69.46) +- (0,0.90)
        (10,80.01) +- (0,0.68)
        (25,88.99) +- (0,0.34)
        (50,92.62) +- (0,0.14)
        (100,94.17) +- (0,0.06)
        (200,94.99) +- (0,0.02)
        (300,95.31) +- (0,0.03)
        (400,95.50) +- (0,0.02)
        (500,95.64) +- (0,0.00)
    };
    \addlegendentry{\unilid}

    \end{axis}
    \end{tikzpicture}}
    \caption{Sample efficiency on WiLI: accuracy against the number of training samples per language, reported as the mean and standard deviation over repeated subsampling runs. \unilid's accuracy is higher than \fasttext's at every subsample size, and its standard deviation is below 1 accuracy point at every subsample size.}
    \label{fig:samples-accuracy}
\end{figure}

\paragraph{Low-resource Regime.}
We evaluate sample efficiency on WiLI by varying the training dataset size.\tiago{Table 1 seems to have quite mixed results, but all the extra experiments are on Wili in which we seem quite better. So maybe justifying why wili could be nice?}
Concretely, we train each model on a subset of the WiLI training set, created by taking \corrrev{$J$} points from each language set. \textit{E.g.}, at \corrrev{$J=5$}, there are a total of $5\times |\languages|$ data points in the entire training set; training subsets are the same for each model at a given value of \corrrev{$J$}.
\cref{fig:samples-accuracy}\tiago{Table 4 currently appears before Table 3. Maybe we should switch their order.} shows system accuracy as a function of \corrrev{$J$}.
Results show that \unilid dramatically outperforms \fasttext\footnote{We tuned \fasttext extensively for this setting but could not improve performance; the best configuration is reported here, with a subset of our sweep of \fasttext hyperparameters in \cref{app:fasttext_epochs}. } in the low-resource regime. 
With 5 samples per language, \unilid reaches 69\% accuracy against 10.5\% for \fasttext; at 25 samples per language \unilid reaches 89\% and at 50 it reaches 93\% (\cref{tab:samples-accuracy}). This is the setting in which labeled data is scarce or expensive to obtain.
We provide the values displayed in \cref{fig:samples-accuracy} in \cref{tab:samples-accuracy} in \cref{app:additional_results}.

\paragraph{Out-of-domain Performance.}
\cref{tab:tatoeba_udhr_comparison} evaluates generalization by training on Wikipedia (WiLI) and testing on short, noisy community-contributed text (Tatoeba) and formal legal documents (UDHR). 
\unilid more than doubles the macro F1 of \fasttext on Tatoeba (0.420 against 0.160), and is 0.017 above it on UDHR (0.866 against 0.849).

\paragraph{Robustness Analysis.}
\cref{tab:length_accuracy} reports accuracy as a function of input length on WiLI. 
\unilid consistently outperforms \fasttext across all length buckets, and the difference is largest for the shortest inputs: 2.3 accuracy points at 101 to 150 characters, against 0.3 points at 1000 characters and above.
This highlights the robustness of \unilid's language-specific segmentation assumptions for low-context settings, which are common in practical LID applications, such as determining the language of social media posts.
We additionally evaluate robustness to character-level corruption on WiLI by stochastically replacing non-whitespace characters at rates of 5\% to 50\%. Full setup details and results can be found in \cref{app:noise_robustness}. In short, at corruption rates up to 10\% the two systems are roughly tied, with \unilid slightly ahead; at 25\% and above \fasttext degrades less (accuracy 0.8925 against 0.8562 at the 25\% rate; \cref{tab:noise_robustness}).

\paragraph{Vocabulary Sensitivity.}
\cref{tab:lid_main} includes three \unilid-X variants that reuse off-the-shelf LLM tokenizer vocabularies (\corrrev{\unilid-Mistral-Nemo, \unilid-DeepSeek3.2, \unilid-Qwen3}); all are \corrrev{within 0.03 macro F1}
of base UniLID on GlotLID-C, with similar behavior for other LLM
tokenizers (\cref{app:additional_results}).
The gap in F1 is not negligible though. Clearly, training the base \unilid tokenizer on broad multilingual data, such that it attains a comprehensive multilingual vocabulary, does lead to better performance---both in-distribution (e.g., GlotLID-C train $
\rightarrow$ GlotLID-C test) and out-of-distribution (GlotLID-C train $\rightarrow$ UDHR, FLORES-200, neither of which appears in the training set; see \cref{tab:lid_main}).  
LLM tokenizers, on the other hand, are optimized for general-purpose language modeling on data distributions that under-represent the low-resource tail of GlotLID-C.  We
accordingly recommend, for production deployment, training the base
tokenizer once on a broad, web-representative multilingual corpus, and
then estimating per-language unigram distributions incrementally as data
for new languages arrives. \unilid-X is intended as a convenience mode for practitioners who wish to reuse an existing LLM tokenizer or to integrate LID into an existing LLM tokenization pipeline \corrrev{without training a second vocabulary}.
We additionally explore the effect of vocabulary size in \cref{app:additional_results}. 
Performance improves modestly with increasing vocabulary size before plateauing, suggesting that UniLID is relatively robust to this choice and does not require extremely large vocabularies to perform well.

\paragraph{Efficiency.} \camrev{\cref{app:efficiency}}
reports inference latency and wall-clock training time for \unilid, \fasttext, and \cld measured on identical hardware. Training UniLID on the full
GlotLID-C corpus (1{,}940 language--script classes) takes 17.8k seconds
end-to-end versus $\sim$163k seconds for the 100-epoch \fasttext
configuration, an approximately $9\times$ reduction despite our
unoptimized research implementation; the comparison is conditioned
on the 100-epoch configuration, which was selected for accuracy (at 1 epoch \fasttext reaches macro F1 0.115 on DSL-ML 2024, against 0.532 for the
configuration used in this paper; \cref{tab:fasttext_epoch_sweep}).
Inference latency varies with the
label set size: on GlotLID-C (1{,}940 labels), \unilid is $1.85\times$ slower
than \fasttext (0.307 vs 0.166 ms/sample) and $1.64\times$ slower than
\cld (0.187 ms/sample); on WiLI (235 labels), UniLID is the second-fastest
of the three (0.155 ms/sample, vs 0.113 for \fasttext and 0.427 for
\cld).\footnote{Latency grows with the size of the label set
(\cref{sec:unilid}), and repeated runs of one configuration vary: the
100k-vocabulary \unilid configuration on WiLI measures 0.155 ms/sample in
\cref{tab:latency_wili} and 0.175 ms/sample in
\cref{tab:vocab_size_efficiency}.} The inference time dependency on label-set size can easily be mitigated in practice: for deployments with restricted target label sets, the generative formulation allows us to remove excluded languages trivially without changing the results. We discuss this next.

\section{Discussion}\label{sec:disc}

\paragraph{Limitations.} A main limitation of \unilid is its reliance on a unigram assumption, which ignores dependencies between neighboring tokens. 
Natural language is inherently contextual, and modeling token interactions is a critical part of most NLP approaches. 
In this respect, \unilid trades expressive power for computational and statistical efficiency. However, this design choice is conceptually aligned with widely adopted baselines such as \fasttext, which treats character $n$-grams as an orderless bag of features and thus likewise ignores neighboring token dependencies.\clara{This could use a bit of work}
The strong empirical performance of both methods suggests that a large fraction of the signal required for practical LID is captured by local subword statistics rather than long-range dependencies.
Further, while \unilid is efficient at inference, its storage and memory requirements scale linearly with the number of modeled languages.  This could prove potentially challenging for scenarios involving thousands of languages under strict latency constraints. Notably, the modularity of our formulation means: (i) the per-language step parallelizes trivially across language partitions; (ii) the generative formulation allows the label set to be subsetted to a deployment's actual target languages without any retraining. Using these characteristics can drastically mitigate any inference and memory bottlenecks.\looseness=-1

\paragraph{Extensions and Future Work.}
Most of calibrated \unilid's remaining errors are confusions between two
well-resourced languages (numbers in \cref{app:protocol}); unseen-token
treatments cannot separate such pairs. The open question is whether a
mechanism that compares language pairs directly can preserve the property
that adding a language needs only that language's own data.

\unilid extends naturally to text classification tasks outside of LID: nothing in the formulation is specific to LID, so estimating a separate unigram distribution per class and scoring by Viterbi segmentation applies to any partition of text with distinct subword statistics, such as domain, register, or source identification.
The unigram assumption of \unilid could also be relaxed to incorporate token dependencies, for example via language-conditional $n$-gram models over token sequences, either built directly into the generative model formulation or learned over text already segmented with the unigram model. This would substantially increase computational complexity, especially in the first case, where segmentation would need to account for token-pair interactions.
Hierarchical decoding would reduce the effective $|\languages|$ by first selecting a script or language family (\eg, via a script detection filter) and then discriminating within it. Sparse representations of the per-language distributions would reduce memory, since those distributions are sparse after EM convergence. Either would mitigate the scaling of inference cost and memory with the number of languages. 

\section{Conclusion}
We introduced \unilid, a simple and efficient approach for language identification. 
In short, to predict a string's language label, we ask: under which language's unigram distribution is this string most likely? \clara{reword so its not overlapping with abstract}
More concretely, we adapt the generative framework of the \unilm tokenization algorithm, creating language-conditional probability distributions that enable segmentation to be treated as a language-specific phenomenon. 
A simple application of Bayes' rule to per-language string likelihoods then provides a posterior over language labels.  
Empirically, \unilid demonstrates performance competitive with widely used LID systems. 
In particular, it exhibits stronger data efficiency and greater robustness in low-resource and fine-grained identification regimes, suggesting it will be a valuable asset for curating high-quality, diverse datasets for next-generation multilingual language models.

\section*{Impact Statement}
This work aims to advance language identification in low-resource and fine-grained dialectal settings. As LID is often an initial step in multilingual data curation pipelines, the performance of these systems can strongly influence which languages and language varieties are supported by large-scale language models.
Consequently, a potential positive impact of this work is to support more inclusive and representative multilingual datasets. 
By improving performance in low-resource and short-text regimes, the proposed method may facilitate the inclusion of languages and language varieties that are underrepresented in current large-scale corpora. The low false positive rate of our method is of particular importance, as data contamination in these settings can have a severe negative impact. 
Thus, we believe this method can contribute towards efforts to reduce the disparity in language model performance on high-resource vs.\ low-resource languages.\looseness=-1

As with many language technologies, better LID capabilities raise dual-use concerns. 
While more accurate identification can facilitate inclusion, it can also be repurposed to filter, restrict, or suppress content associated with specific linguistic communities. In addition, fine-grained dialect identification could be misused for profiling or monitoring individuals or groups based on linguistic characteristics. These risks are not unique to the proposed method but are common to LID and related text classification technologies more broadly. We encourage practitioners to apply LID systems in ways that are consistent with ethical data collection and responsible use of language technologies.

\Clara{
Experimental Results to Add (in order of importance; last ones are not critical)
\begin{itemize}
    \item WiLI results (I believe we already have these, they just need to be added) \done
    \item Results with open-source LLM base tokenizers (as additional rows in \cref{tab:lid_main})  \done
    \item \glotlid results on UDHR (also as additional rows in \cref{tab:lid_main}) \done
    \item Domain shift results (train on WiLI, evaluate on something like UDHR or Tatoeba) (separate table) \done
    \item Ablations on WiLI with different vocab sizes and a few different open source LLM tokenizers \done
    \item Some sort of std dev or std error estimates...
    \item Prompting an LLM to label the language
    \item Results with per-language trained unigramLM tokenizers 
    \item using marginal probabilities instead of only probability from max-probability segmentation
\end{itemize}}

\bibliography{custom}

@inproceedings{chifu-etal-2024-vardial,
    title = "{V}ar{D}ial Evaluation Campaign 2024: {C}ommonsense Reasoning in Dialects and Multi-Label Similar Language Identification",
    author = "Chifu, Adrian-Gabriel  and
      Glava{\v{s}}, Goran  and
      Ionescu, Radu Tudor  and
      Ljube{\v{s}}i{\'c}, Nikola  and
      Mileti{\'c}, Aleksandra  and
      Mileti{\'c}, Filip  and
      Scherrer, Yves  and
      Vuli{\'c}, Ivan",
    editor = {Scherrer, Yves  and
      Jauhiainen, Tommi  and
      Ljube{\v{s}}i{\'c}, Nikola  and
      Zampieri, Marcos  and
      Nakov, Preslav  and
      Tiedemann, J{\"o}rg},
    booktitle = "Proceedings of the Eleventh Workshop on NLP for Similar Languages, Varieties, and Dialects (VarDial 2024)",
    month = jun,
    year = "2024",
    address = "Mexico City, Mexico",
    publisher = "Association for Computational Linguistics",
    url = "https://aclanthology.org/2024.vardial-1.1/",
    doi = "10.18653/v1/2024.vardial-1.1",
    pages = "1--15"
}

@misc{meister2025unigramlm,
  title   = {{UnigramLM}: {An} Attempt at Writing The Missing Manual},
  author  = {Meister, Clara},
  year    = {2025},
  url     = {https://cimeister.github.io/blog/unigramlm/},
  note    = {Blog post}
}

@inproceedings{foroutan2025conlid,
    title = "{C}on{LID}: Supervised Contrastive Learning for Low-Resource Language Identification",
    author = "Foroutan, Negar  and
      Saydaliev, Jakhongir  and
      Kim, Grace  and
      Bosselut, Antoine",
    editor = "Demberg, Vera  and
      Inui, Kentaro  and
      Marquez, Llu{\'i}s",
    booktitle = "Proceedings of the 19th Conference of the {E}uropean Chapter of the {A}ssociation for {C}omputational {L}inguistics (Volume 1: Long Papers)",
    month = mar,
    year = "2026",
    address = "Rabat, Morocco",
    publisher = "Association for Computational Linguistics",
    url = "https://aclanthology.org/2026.eacl-long.315/",
    doi = "10.18653/v1/2026.eacl-long.315",
    pages = "6693--6708",
    ISBN = "979-8-89176-380-7"
}

@inproceedings{kudo-2018-subword,
    title = "Subword Regularization: Improving Neural Network Translation Models with Multiple Subword Candidates",
    author = "Kudo, Taku",
    editor = "Gurevych, Iryna  and
      Miyao, Yusuke",
    booktitle = "Proceedings of the 56th Annual Meeting of the Association for Computational Linguistics (Volume 1: Long Papers)",
    month = jul,
    year = "2018",
    address = "Melbourne, Australia",
    publisher = "Association for Computational Linguistics",
    url = "https://aclanthology.org/P18-1007/",
    doi = "10.18653/v1/P18-1007",
    pages = "66--75"
}

@inproceedings{lui-baldwin-2012-langid,
    title = "langid.py: An Off-the-shelf Language Identification Tool",
    author = "Lui, Marco  and
      Baldwin, Timothy",
    editor = "Zhang, Min",
    booktitle = "Proceedings of the {ACL} 2012 System Demonstrations",
    month = jul,
    year = "2012",
    address = "Jeju Island, Korea",
    publisher = "Association for Computational Linguistics",
    url = "https://aclanthology.org/P12-3005/",
    pages = "25--30"
}

@inproceedings{joulin-etal-2017-bag,
    title = "Bag of Tricks for Efficient Text Classification",
    author = "Joulin, Armand  and
      Grave, Edouard  and
      Bojanowski, Piotr  and
      Mikolov, Tomas",
    editor = "Lapata, Mirella  and
      Blunsom, Phil  and
      Koller, Alexander",
    booktitle = "Proceedings of the 15th Conference of the {E}uropean Chapter of the Association for Computational Linguistics: Volume 2, Short Papers",
    month = apr,
    year = "2017",
    address = "Valencia, Spain",
    publisher = "Association for Computational Linguistics",
    url = "https://aclanthology.org/E17-2068/",
    pages = "427--431"
}

@article{Thoma2018TheWB,
  title={The {WiLI} benchmark dataset for written language identification},
  author={Martin Thoma},
  journal={CoRR},
  year={2018},
  volume={abs/1801.07779},
  url={https://api.semanticscholar.org/CorpusID:3495417}
}

@article{kneser,
  title={On the Estimation of Small Probabilities by Leaving-One-Out},
  author={Ney, Hermann and Essen, Ute and Kneser, Reinhard},
  journal={IEEE Transactions on Pattern Analysis and Machine Intelligence},
  volume={17},
  number={12},
  pages={1202--1212},
  year={1995},
  publisher = {IEEE Computer Society},
  url = {https://ieeexplore.ieee.org/document/476512},
address = {USA},
}

@inproceedings{baldwin-lui-2010-language,
    title = "Language Identification: The Long and the Short of the Matter",
    author = "Baldwin, Timothy  and
      Lui, Marco",
    editor = "Kaplan, Ron  and
      Burstein, Jill  and
      Harper, Mary  and
      Penn, Gerald",
    booktitle = "Human Language Technologies: The 2010 Annual Conference of the North {A}merican Chapter of the Association for Computational Linguistics",
    month = jun,
    year = "2010",
    address = "Los Angeles, California",
    publisher = "Association for Computational Linguistics",
    url = "https://aclanthology.org/N10-1027/",
    pages = "229--237"
}

@article{ren-etal-2022-effective,
    title = "Effective Approaches to Neural Query Language Identification",
    author = "Ren, Xingzhang  and
      Yang, Baosong  and
      Liu, Dayiheng  and
      Zhang, Haibo  and
      Lv, Xiaoyu  and
      Yao, Liang  and
      Xie, Jun",
    journal = "Computational Linguistics",
    volume = "48",
    number = "4",
    month = dec,
    year = "2022",
    address = "Cambridge, MA",
    publisher = "MIT Press",
    url = "https://aclanthology.org/2022.cl-4.14/",
    doi = "10.1162/coli_a_00451",
    pages = "887--906"
}

@inproceedings{ng2001discriminative,
  title     = {On discriminative vs. generative classifiers: A comparison of logistic regression and naive {B}ayes},
  author    = {Ng, Andrew Y. and Jordan, Michael I.},
  booktitle = {Advances in Neural Information Processing Systems},
  editor    = {Dietterich, T. and Becker, S. and Ghahramani, Z.},
  volume    = {14},
  year      = {2001},
  publisher = {MIT Press}
}

@inproceedings{kudo-richardson-2018-sentencepiece,
    title = "{S}entence{P}iece: A simple and language independent subword tokenizer and detokenizer for Neural Text Processing",
    author = "Kudo, Taku  and
      Richardson, John",
    editor = "Blanco, Eduardo  and
      Lu, Wei",
    booktitle = "Proceedings of the 2018 Conference on Empirical Methods in Natural Language Processing: System Demonstrations",
    month = nov,
    year = "2018",
    address = "Brussels, Belgium",
    publisher = "Association for Computational Linguistics",
    url = "https://aclanthology.org/D18-2012/",
    doi = "10.18653/v1/D18-2012",
    pages = "66--71"
}

@inproceedings{tiedemann-2020-tatoeba,
    title = "The Tatoeba Translation Challenge {--} Realistic Data Sets for Low Resource and Multilingual {MT}",
    author = {Tiedemann, J{\"o}rg},
    editor = {Barrault, Lo{\"i}c  and
      Bojar, Ond{\v{r}}ej  and
      Bougares, Fethi  and
      Chatterjee, Rajen  and
      Costa-juss{\`a}, Marta R.  and
      Federmann, Christian  and
      Fishel, Mark  and
      Fraser, Alexander  and
      Graham, Yvette  and
      Guzman, Paco  and
      Haddow, Barry  and
      Huck, Matthias  and
      Yepes, Antonio Jimeno  and
      Koehn, Philipp  and
      Martins, Andr{\'e}  and
      Morishita, Makoto  and
      Monz, Christof  and
      Nagata, Masaaki  and
      Nakazawa, Toshiaki  and
      Negri, Matteo},
    booktitle = "Proceedings of the Fifth Conference on Machine Translation",
    month = nov,
    year = "2020",
    address = "Online",
    publisher = "Association for Computational Linguistics",
    url = "https://aclanthology.org/2020.wmt-1.139/",
    pages = "1174--1182"
}

@inproceedings{gaman-etal-2020-report,
    title = "A Report on the {V}ar{D}ial Evaluation Campaign 2020",
    author = "Gaman, Mihaela  and
      Hovy, Dirk  and
      Ionescu, Radu Tudor  and
      Jauhiainen, Heidi  and
      Jauhiainen, Tommi  and
      Lind{\'e}n, Krister  and
      Ljube{\v{s}}i{\'c}, Nikola  and
      Partanen, Niko  and
      Purschke, Christoph  and
      Scherrer, Yves  and
      Zampieri, Marcos",
    editor = {Zampieri, Marcos  and
      Nakov, Preslav  and
      Ljube{\v{s}}i{\'c}, Nikola  and
      Tiedemann, J{\"o}rg  and
      Scherrer, Yves},
    booktitle = "Proceedings of the 7th Workshop on NLP for Similar Languages, Varieties and Dialects",
    month = dec,
    year = "2020",
    address = "Barcelona, Spain (Online)",
    publisher = "International Committee on Computational Linguistics (ICCL)",
    url = "https://aclanthology.org/2020.vardial-1.1/",
    pages = "1--14"
}

@inproceedings{caswell-etal-2020-language,
    title = "Language {ID} in the Wild: Unexpected Challenges on the Path to a Thousand-Language Web Text Corpus",
    author = "Caswell, Isaac  and
      Breiner, Theresa  and
      van Esch, Daan  and
      Bapna, Ankur",
    editor = "Scott, Donia  and
      Bel, Nuria  and
      Zong, Chengqing",
    booktitle = "Proceedings of the 28th International Conference on Computational Linguistics",
    month = dec,
    year = "2020",
    address = "Barcelona, Spain (Online)",
    publisher = "International Committee on Computational Linguistics",
    url = "https://aclanthology.org/2020.coling-main.579/",
    doi = "10.18653/v1/2020.coling-main.579",
    pages = "6588--6608"
}

@article{mcnamee_solved,
author = {McNamee, Paul},
title = {Language identification: a solved problem suitable for undergraduate instruction},
year = {2005},
issue_date = {February 2005},
publisher = {Consortium for Computing Sciences in Colleges},
address = {Evansville, IN, USA},
volume = {20},
number = {3},
issn = {1937-4771},
journal = {J. Comput. Sci. Coll.},
month = feb,
pages = {94–101},
numpages = {8}
}

@inproceedings{Abouelhoda2002TheES,
  title={The Enhanced Suffix Array and Its Applications to Genome Analysis},
  author={Mohamed Ibrahim Abouelhoda and Stefan Kurtz and Enno Ohlebusch},
  booktitle={Workshop on Algorithms in Bioinformatics},
  year={2002},
  url={https://api.semanticscholar.org/CorpusID:1161031}
}

@inproceedings{cavnar01,
    author = {Cavnar, William B. and Trenkle, John M.},
    year = {1994},
    title = {N-Gram-Based Text Categorization},
    booktitle = {Proceedings of SDAIR-94, 3rd Annual Symposium on Document Analysis and Information Retrieval},
    address = {Las Vegas, US},
    pages = {161--175}
}

@article{nllb2022,
  author    = {{NLLB Team} and Costa-juss{\`a}, Marta R. and Cross, James and {\c{C}}elebi, Onur and Elbayad, Maha and Heafield, Kenneth and Heffernan, Kevin and Kalbassi, Elahe and Lam, Janice and Licht, Daniel and Maillard, Jean and Sun, Anna and Wang, Skyler and Wenzek, Guillaume and Youngblood, Al and Akula, Bapi and Barrault, Loic and Gonzalez, Gabriel Mejia and Hansanti, Prangthip and Hoffman, John and Jarrett, Semarley and Sadagopan, Kaushik Ram and Rowe, Dirk and Spruit, Shannon and Tran, Chau and Andrews, Pierre and Ayan, Necip Fazil and Bhosale, Shruti and Edunov, Sergey and Fan, Angela and Gao, Cynthia and Goswami, Vedanuj and Guzm{\'a}n, Francisco and Koehn, Philipp and Mourachko, Alexandre and Ropers, Christophe and Saleem, Safiyyah and Schwenk, Holger and Wang, Jeff},
  title     = {No Language Left Behind: Scaling Human-Centered Machine Translation},
  journal   = {CoRR},
  volume    = {abs/2207.04672},
  year      = {2022},
  url       = {https://arxiv.org/abs/2207.04672}
}

@InProceedings{Vatanen10LREC,
    author =    {Tommi Vatanen and Jaakko J. V{\"{a}}yrynen and Sami Virpioja},
    booktitle = {Proceedings of the Seventh conference on International Language Resources and Evaluation (LREC'10)},
    pages =     {3423--3430},
    publisher = {European Language Resources Association (ELRA)},
    title =     {Language identification of short text segments with n-gram models},
    year =      {2010},
}

@inproceedings{sennrich-etal-2016-neural,
    title = "Neural Machine Translation of Rare Words with Subword Units",
    author = "Sennrich, Rico  and
      Haddow, Barry  and
      Birch, Alexandra",
    editor = "Erk, Katrin  and
      Smith, Noah A.",
    booktitle = "Proceedings of the 54th Annual Meeting of the Association for Computational Linguistics (Volume 1: Long Papers)",
    month = aug,
    year = "2016",
    address = "Berlin, Germany",
    publisher = "Association for Computational Linguistics",
    url = "https://aclanthology.org/P16-1162/",
    doi = "10.18653/v1/P16-1162",
    pages = "1715--1725"
}

@article{lid_survey,
author = {Jauhiainen, Tommi and Lui, Marco and Zampieri, Marcos and Baldwin, Timothy and Lindén, Krister},
year = {2019},
month = {08},
pages = {},
title = {Automatic Language Identification in Texts: A Survey},
volume = {65},
journal = {Journal of Artificial Intelligence Research},
doi = {10.1613/jair.1.11675}
}

@inproceedings{burchell-etal-2023-open,
    title = "An Open Dataset and Model for Language Identification",
    author = "Burchell, Laurie  and
      Birch, Alexandra  and
      Bogoychev, Nikolay  and
      Heafield, Kenneth",
    editor = "Rogers, Anna  and
      Boyd-Graber, Jordan  and
      Okazaki, Naoaki",
    booktitle = "Proceedings of the 61st Annual Meeting of the Association for Computational Linguistics (Volume 2: Short Papers)",
    month = jul,
    year = "2023",
    address = "Toronto, Canada",
    publisher = "Association for Computational Linguistics",
    url = "https://aclanthology.org/2023.acl-short.75/",
    doi = "10.18653/v1/2023.acl-short.75",
    pages = "865--879"
}

@inproceedings{goot-2025-identifying,
    title = "Identifying Open Challenges in Language Identification",
    author = "Goot, Rob Van Der",
    editor = "Che, Wanxiang  and
      Nabende, Joyce  and
      Shutova, Ekaterina  and
      Pilehvar, Mohammad Taher",
    booktitle = "Proceedings of the 63rd Annual Meeting of the Association for Computational Linguistics (Volume 1: Long Papers)",
    month = jul,
    year = "2025",
    address = "Vienna, Austria",
    publisher = "Association for Computational Linguistics",
    url = "https://aclanthology.org/2025.acl-long.891/",
    doi = "10.18653/v1/2025.acl-long.891",
    pages = "18207--18227",
    ISBN = "979-8-89176-251-0"
}

@techreport{dunning1994statistical,
  author       = {Dunning, Ted},
  title        = {Statistical Identification of Language},
  institution  = {New Mexico State University, Computer Research Laboratory},
  year         = {1994},
  number       = {MCCS-94-273},
  type         = {Technical Report}
}

@inproceedings{zhang_cnn_2015,
author = {Zhang, Xiang and Zhao, Junbo and LeCun, Yann},
title = {Character-level convolutional networks for text classification},
year = {2015},
publisher = {MIT Press},
address = {Cambridge, MA, USA},
booktitle = {Proceedings of the 29th International Conference on Neural Information Processing Systems - Volume 1},
pages = {649–657},
numpages = {9},
location = {Montreal, Canada},
series = {NIPS'15}
}

@inproceedings{kargaran-etal-2023-glotlid,
    title = "{G}lot{LID}: Language Identification for Low-Resource Languages",
    author = "Kargaran, Amir Hossein  and
      Imani, Ayyoob  and
      Yvon, Fran{\c{c}}ois  and
      Schuetze, Hinrich",
    editor = "Bouamor, Houda  and
      Pino, Juan  and
      Bali, Kalika",
    booktitle = "Findings of the Association for Computational Linguistics: EMNLP 2023",
    month = dec,
    year = "2023",
    address = "Singapore",
    publisher = "Association for Computational Linguistics",
    url = "https://aclanthology.org/2023.findings-emnlp.410/",
    doi = "10.18653/v1/2023.findings-emnlp.410",
    pages = "6155--6218"
}

@inproceedings{land2025piecesdoesunigramtokenization,
    title = "Which Pieces Does Unigram Tokenization Really Need?",
    author = "Land, Sander  and
      Pinter, Yuval",
    editor = "Liakata, Maria  and
      Moreira, Viviane P.  and
      Zhang, Jiajun  and
      Jurgens, David",
    booktitle = "Findings of the {A}ssociation for {C}omputational {L}inguistics: {ACL} 2026",
    month = jul,
    year = "2026",
    address = "San Diego, California, United States",
    publisher = "Association for Computational Linguistics",
    url = "https://aclanthology.org/2026.findings-acl.316/",
    doi = "10.18653/v1/2026.findings-acl.316",
    pages = "6351--6360",
    ISBN = "979-8-89176-395-1"
}

@article{ojo2025diversbenchevaluatinglanguageidentification,
      title={{DIVERS}-{B}ench: Evaluating Language Identification Across Domain Shifts and Code-Switching}, 
      author={Jessica Ojo and Zina Kamel and David Ifeoluwa Adelani},
      year={2025},
      eprint={2509.17768},
      journal={CoRR},
    volume={abs/2509.17768},
      url={https://arxiv.org/abs/2509.17768}, 
}

@article{kreutzer-etal-2022-quality,
    title = "Quality at a Glance: An Audit of Web-Crawled Multilingual Datasets",
    author = {Kreutzer, Julia  and
      Caswell, Isaac  and
      Wang, Lisa  and
      Wahab, Ahsan  and
      van Esch, Daan  and
      Ulzii-Orshikh, Nasanbayar  and
      Tapo, Allahsera  and
      Subramani, Nishant  and
      Sokolov, Artem  and
      Sikasote, Claytone  and
      Setyawan, Monang  and
      Sarin, Supheakmungkol  and
      Samb, Sokhar  and
      Sagot, Beno{\^i}t  and
      Rivera, Clara  and
      Rios, Annette  and
      Papadimitriou, Isabel  and
      Osei, Salomey  and
      Suarez, Pedro Ortiz  and
      Orife, Iroro  and
      Ogueji, Kelechi  and
      Rubungo, Andre Niyongabo  and
      Nguyen, Toan Q.  and
      M{\"u}ller, Mathias  and
      M{\"u}ller, Andr{\'e}  and
      Muhammad, Shamsuddeen Hassan  and
      Muhammad, Nanda  and
      Mnyakeni, Ayanda  and
      Mirzakhalov, Jamshidbek  and
      Matangira, Tapiwanashe  and
      Leong, Colin  and
      Lawson, Nze  and
      Kudugunta, Sneha  and
      Jernite, Yacine  and
      Jenny, Mathias  and
      Firat, Orhan  and
      Dossou, Bonaventure F. P.  and
      Dlamini, Sakhile  and
      de Silva, Nisansa  and
      {\c{C}}abuk Ball{\i}, Sakine  and
      Biderman, Stella  and
      Battisti, Alessia  and
      Baruwa, Ahmed  and
      Bapna, Ankur  and
      Baljekar, Pallavi  and
      Azime, Israel Abebe  and
      Awokoya, Ayodele  and
      Ataman, Duygu  and
      Ahia, Orevaoghene  and
      Ahia, Oghenefego  and
      Agrawal, Sweta  and
      Adeyemi, Mofetoluwa},
    editor = "Roark, Brian  and
      Nenkova, Ani",
    journal = "Transactions of the Association for Computational Linguistics",
    volume = "10",
    year = "2022",
    address = "Cambridge, MA",
    publisher = "MIT Press",
    url = "https://aclanthology.org/2022.tacl-1.4/",
    doi = "10.1162/tacl_a_00447",
    pages = "50--72"
}

@inproceedings{chen-etal-2024-fumbling,
    title = "Fumbling in {B}abel: An Investigation into {C}hat{GPT}{'}s Language Identification Ability",
    author = "Chen, Wei-Rui  and
      Adebara, Ife  and
      Doan, Khai  and
      Liao, Qisheng  and
      Abdul-Mageed, Muhammad",
    editor = "Duh, Kevin  and
      Gomez, Helena  and
      Bethard, Steven",
    booktitle = "Findings of the Association for Computational Linguistics: NAACL 2024",
    month = jun,
    year = "2024",
    address = "Mexico City, Mexico",
    publisher = "Association for Computational Linguistics",
    url = "https://aclanthology.org/2024.findings-naacl.274/",
    doi = "10.18653/v1/2024.findings-naacl.274",
    pages = "4387--4413"
}

@inproceedings{suarez-etal-2026-commonlid,
    title = "{C}ommon{LID}: Re-evaluating State-of-the-Art Language Identification Performance on Web Data",
    author = "Suarez, Pedro Ortiz  and
      Burchell, Laurie  and
      Arnett, Catherine  and
      Mosquera, Rafael  and
      Monsalve, Sara Hincapi{\'e}  and
      Vaughan, Thom  and
      Stewart, Damian  and
      Ostendorff, Malte  and
      Abdulmumin, Idris  and
      Marivate, Vukosi  and
      Muhammad, Shamsuddeen Hassan  and
      Tonja, Atnafu Lambebo  and
      Al-Khalifa, Hend  and
      Hammouda, Nadia Ghezaiel  and
      Otiende, Verrah Akinyi  and
      Wong, Tack Hwa  and
      Saydaliev, Jakhongir  and
      Nobakhtian, Melika  and
      Habibi, Muhammad Ravi Shulthan  and
      Kranti, Chalamalasetti  and
      Muchemi, Carol  and
      Nguyen, Khang  and
      Adam, Faisal Muhammad  and
      Salim, Luis Frentzen  and
      Alqifari, Reem  and
      Amol, Cynthia Jayne  and
      Imperial, Joseph Marvin  and
      Kesen, Ilker  and
      Mustafid, Ahmad  and
      Stepachev, Pavel  and
      Choshen, Leshem  and
      Anugraha, David  and
      Nayel, Hamada  and
      Yimam, Seid Muhie  and
      Alexandra Putra, Vallerie  and
      Nguyen, My Chiffon  and
      Wasi, Azmine Toushik  and
      Vadithya, Gouthami  and
      van der Goot, Rob  and
      C{'}horr, Lanwenn ar  and
      Dua, Karan  and
      Yates, Andrew  and
      Bangera, Mithil  and
      Bangera, Yeshil  and
      Patel, Hitesh Laxmichand  and
      Okabe, Shu  and
      Ilasariya, Fenal Ashokbhai  and
      Gaynullin, Dmitry  and
      Winata, Genta Indra  and
      Li, Yiyuan  and
      Mart{\'i}nez, Juan Pablo  and
      Agarwal, Amit  and
      Hanif, Ikhlasul Akmal  and
      Ahmad, Raia Abu  and
      Adenuga, Esther  and
      Tjiaranata, Filbert Aurelian  and
      Buaphet, Weerayut  and
      Anugraha, Michael  and
      Vajjala, Sowmya  and
      Rice, Benjamin L  and
      Amirudin, Azril Hafizi  and
      Alabi, Jesujoba Oluwadara  and
      Panda, Srikant  and
      Toughrai, Yassine  and
      Kyomuhendo, Bruhan  and
      Ruffinelli, Daniel  and
      Akshata  and
      Goul{\~a}o, Manuel  and
      Zhou, Ej  and
      Ramirez, Ingrid Gabriela Franco  and
      Aggazzotti, Cristina  and
      Dobler, Konstantin  and
      Kevin, Jun  and
      Pag{\`e}s, Quentin  and
      Andrews, Nicholas  and
      Ibrahim, Nuhu  and
      Ruckdeschel, Mattes  and
      Keleg, Amr  and
      Zhang, Mike  and
      Muziri, Casper Rufaro  and
      Samuel, Saron  and
      Takeshita, Sotaro  and
      Kerdthaisong, Kun  and
      Foppiano, Luca  and
      Dent, Rasul  and
      Green, Tommaso  and
      Wali, Ahmad Mustapha  and
      Makaaka, Kamohelo  and
      Feliren, Vicky  and
      Idris, Inshirah  and
      Celikkanat, Hande  and
      Abubakar, Abdulhamid  and
      Maillard, Jean  and
      Sagot, Beno{\^i}t  and
      Cl{\'e}rice, Thibault  and
      Murray, Kenton  and
      Luger, Sarah K. K.",
    editor = "Liakata, Maria  and
      Moreira, Viviane P.  and
      Zhang, Jiajun  and
      Jurgens, David",
    booktitle = "Proceedings of the 64th Annual Meeting of the {A}ssociation for {C}omputational {L}inguistics (Volume 1: Long Papers)",
    month = jul,
    year = "2026",
    address = "San Diego, California, United States",
    publisher = "Association for Computational Linguistics",
    url = "https://aclanthology.org/2026.acl-long.1527/",
    doi = "10.18653/v1/2026.acl-long.1527",
    pages = "33063--33080",
    ISBN = "979-8-89176-390-6"
}

@article{dempster1977maximum,
 ISSN = {00359246},
 URL = {http://www.jstor.org/stable/2984875},
 author = {A. P. Dempster and N. M. Laird and D. B. Rubin},
 journal = {Journal of the Royal Statistical Society. Series B (Methodological)},
 number = {1},
 pages = {1--38},
 publisher = {[Royal Statistical Society, Oxford University Press]},
 title = {Maximum Likelihood from Incomplete Data via the EM Algorithm},
 urldate = {2026-09-09},
 volume = {39},
 year = {1977}
}

@inproceedings{belinkov-glass-2016-character,
    title = "A Character-level Convolutional Neural Network for Distinguishing Similar Languages and Dialects",
    author = "Belinkov, Yonatan  and
      Glass, James",
    editor = {Nakov, Preslav  and
      Zampieri, Marcos  and
      Tan, Liling  and
      Ljube{\v{s}}i{\'c}, Nikola  and
      Tiedemann, J{\"o}rg  and
      Malmasi, Shervin},
    booktitle = "Proceedings of the Third Workshop on {NLP} for Similar Languages, Varieties and Dialects ({V}ar{D}ial3)",
    month = dec,
    year = "2016",
    address = "Osaka, Japan",
    publisher = "The COLING 2016 Organizing Committee",
    url = "https://aclanthology.org/W16-4819/",
    pages = "145--152"
}

@inproceedings{kocmi-bojar-2017-lanidenn,
    title = "{L}anide{NN}: Multilingual Language Identification on Character Window",
    author = "Kocmi, Tom  and
      Bojar, Ond{\v{r}}ej",
    editor = "Lapata, Mirella  and
      Blunsom, Phil  and
      Koller, Alexander",
    booktitle = "Proceedings of the 15th Conference of the {E}uropean Chapter of the Association for Computational Linguistics: Volume 1, Long Papers",
    month = apr,
    year = "2017",
    address = "Valencia, Spain",
    publisher = "Association for Computational Linguistics",
    url = "https://aclanthology.org/E17-1087/",
    pages = "927--936"
}
\bibliographystyle{icml2026}

\newpage
\clearpage
\appendix

\section{Additional Dataset Details}
\label{app:datasets}

\paragraph{GlotLID-C.}
GlotLID-C \cite{kargaran-etal-2023-glotlid} is a massive open-source benchmark designed to address the long-tail of LID. 
It covers $1940$ different language--script classes,\footnote{This is the number of unique labels in the officially maintained corpus on HuggingFace, which is what we employed: \href{https://huggingface.co/datasets/cis-lmu/glotlid-corpus}{datasets/cis-lmu/glotlid-corpus}.} significantly expanding the label space compared to traditional benchmarks.  
The dataset places a strong emphasis on low-resource and closely related language varieties. 
Train/test splits were not released by the dataset authors; we create our own splits following the pipeline of \citet{foroutan2025conlid}: for each language, we split the data into 85\%/15\% for training/testing. 
We then subset the training set for each language to be at most $100k$ samples. Finally, we check explicitly for contamination and remove all occurrences of data from our eval only sets (described next) that we find in the training set. The resulting training set contains approximately $60M$ samples and test set contains approximately $45M$ samples of 1940 language--script combinations. The \fasttext baseline is trained on the same training set as \unilid's per-language EM runs.

\paragraph{UDHR \citep[eval only;][]{Vatanen10LREC}.}
The Universal Declaration of Human Rights (UDHR)\footnote{\url{https://www.un.org/en/about-us/universal-declaration-of-human-rights}} dataset contains parallel text across hundreds of languages. 
Because the semantic content is identical across all samples, UDHR is often used as a control dataset in LID; the parallel texts allow us to control for content difficulty when assessing per language performance. 
In LID, UDHR is only used as a test set, and its domain differs from most LID training sets, making it well-suited for testing generalization robustness.   

\paragraph{FLORES-200 \citep[eval only;][]{nllb2022}.} 
FLORES-200  is another benchmark consisting of parallel texts. It covers 204 languages. 
Unlike crowdsourced datasets, FLORES-200 consists of professional translations of sentences extracted from Wikimedia projects (WikiNews, WikiVoyage, and WikiBooks). 
It thus provides a controlled test set for assessing performance on low-resource languages. We do not train on the FLORES-200 dataset; we evaluate on the \texttt{devtest} portion.

\paragraph{DSL-ML 2024 \cite{chifu-etal-2024-vardial}.}
To evaluate performance on closely related languages, we use the DSL-ML 2024 shared task dataset. This benchmark focuses on fine-grained dialect identification with 14 labels covering regional variants of five major languages  (English, French, Portuguese, Spanish, and South Slavic). 
It explicitly evaluates a domain in which lexical overlap is high and differentiating factors between dialects are nuanced. 
We use the train and dev splits released by the shared task, training on the train set and evaluating on the dev set. 
\begin{table}[h]
    \centering
    \begin{tabular}{lrr}
        \toprule
        \textbf{Dialect} & \textbf{Train Samples} & \textbf{Test Size} \\
        \midrule
        FR-BE & 122,369 & 7,543 \\
        FR-CH & 116,490 & 1,027 \\
        FR-FR & 84,402  & 6,351 \\
        FR-CA & 19,041  & 2,169 \\
        ES-ES & 2,616   & 631   \\
        ES-AR & 1,982   & 358   \\
        PT-PT & 1,331   & 356   \\
        PT-BR & 2,556   & 635   \\
        EN-US & 1,342   & 372   \\
        EN-GB & 1,028   & 227   \\
        SR    & 236     & 86    \\
        HR    & 53      & 16    \\
        BS    & 45      & 16    \\
        ME    & 34      & 4     \\
        \bottomrule
    \end{tabular}
    \caption{Train and test set sizes by dialect on DSL-ML 2024. We provide these details to help explain the results observed in \cref{tab:per_language_f1}.}
    \label{tab:dialect_stats}
\end{table}
 
\paragraph{WiLI-2018 \citep{Thoma2018TheWB}.} The Wikipedia Language Identification benchmark  contains 235,000 paragraphs balanced equally across 235 languages, all derived from Wikipedia.
We use WiLI-2018 to study data efficiency, base tokenizer choice, and the impact of input sequence length on classification accuracy, as it provides sufficient depth (1,000 samples per language, 500 train/500 test) \corrrev{to support ablations with 500 test samples per language}.
We use the official train/test splits.

\paragraph{Tatoeba (eval only) \cite{tiedemann-2020-tatoeba}.} Tatoeba is a large, community-curated multilingual corpus of short, user-contributed sentences covering hundreds of languages and language varieties.
 The dataset is characterized by wide linguistic diversity, informal style, and substantial variation in sentence length and orthography, making it well suited for evaluating LID systems under realistic, noisy conditions. We use Tatoeba to assess robustness in low-resource and cross-lingual settings, where training data are limited and closely related languages frequently co-occur. As we use this dataset for testing out-of-domain robustness, we do not train on any portion of it. Models are evaluated on the entire dataset for the subset of languages that are in their label set. For evaluations with models trained on WiLI, this results in an evaluation on approximately $~12M$ samples across 201 languages.

\paragraph{CommonLID (eval only).} CommonLID is a web-domain
evaluation set of 373{,}230 lines drawn from Common Crawl, labeled with ISO 639-3 tags, including macrolanguage tags such as Arabic, Malay,
and Chinese \corrrev{\citep{suarez-etal-2026-commonlid}}. Because its labels carry no script and
include macrolanguages while \unilid predicts language--script labels, we
map each prediction to its ISO 639-3 code and score with
macrolanguage-aware accuracy: a prediction is correct if its code equals
the gold tag or is a member of the gold macrolanguage
(\cref{tab:commonlid}). We use CommonLID only as an out-of-domain
evaluation set for calibrated \unilid; no training or selection used it.

\section{Additional Ablation Results}
\label{app:additional_results}
\begin{table}[!h]
\centering
\small
\setlength{\tabcolsep}{6pt}
\renewcommand{\arraystretch}{1.15}
\caption{Accuracy (mean $\pm$ std) as a function of the number of training samples per language on WiLI.}
\adjustbox{max width=\linewidth}{
\begin{tabular}{c c c}
\hline
\textbf{Samples / Language} 
& \textbf{\unilid Acc. (\%) $\uparrow$} 
& \textbf{\fasttext Acc. (\%) $\uparrow$} \\
\hline
5   & \textbf{69.46 $\pm$ 0.90} & 10.53 $\pm$ 0.43 \\
10  & \textbf{80.01 $\pm$ 0.68} & 0.85 $\pm$ 0.00 \\
25  & \textbf{88.99 $\pm$ 0.34} & 13.25 $\pm$ 0.53 \\
50  & \textbf{92.62 $\pm$ 0.14} & 67.79 $\pm$ 0.52 \\
100 & \textbf{94.17 $\pm$ 0.06} & 85.70 $\pm$ 0.40 \\
200 & \textbf{94.99 $\pm$ 0.02} & 91.70 $\pm$ 0.04 \\
300 & \textbf{95.31 $\pm$ 0.03} & 93.20 $\pm$ 0.04 \\
400 & \textbf{95.50 $\pm$ 0.02} & 94.03 $\pm$ 0.03 \\
500 & \textbf{\corrrev{95.64} $\pm$ 0.00} & 94.55 $\pm$ 0.06 \\
\hline
\end{tabular}}

\label{tab:samples-accuracy}
\end{table}
 \label{sec:ablations}
\begin{table}[!h]
\centering
\small
\setlength{\tabcolsep}{6pt}
\renewcommand{\arraystretch}{1.15}
\caption{Effect of base tokenizer vocabulary size on \unilid performance and inference efficiency evaluated on WiLI. \corrrev{Macro F1 and macro FPR come from base tokenizers retrained for this table; at $100$k the retrained vocabulary contains the same $100{,}000$ tokens as the published one, in a different order at 132 of the positions. The latency and throughput columns are the original measurements and were not re-measured on the retrained models.}}
\adjustbox{max width=\linewidth}{
\begin{tabular}{c c c c c}
\hline
\textbf{Vocab Size} &
\textbf{Macro F1}$\uparrow$ &
\textbf{Macro FPR}$\downarrow$ &
\textbf{Latency (ms)}$\downarrow$ &
\textbf{Samples/s}$\uparrow$ \\
\hline
10k   & \corrrev{0.944} & \corrrev{2.556e-4} & 0.113 & 8891.99 \\
20k   & \corrrev{0.950} & \corrrev{2.303e-4} & 0.119 & 8421.68 \\
50k   & \corrrev{0.957} & \corrrev{2.015e-4} & 0.147 & 6818.08 \\
100k  & \corrrev{0.960} & \corrrev{1.863e-4} & 0.175 & 5717.68 \\
200k  & \textbf{\corrrev{0.9606}} & \textbf{\corrrev{1.8418e-4}} & 0.2328 & 4296.45 \\
\hline
\end{tabular}}
\label{tab:vocab_size_efficiency}
\end{table}

\begin{table}[h]
\centering
\small
\setlength{\tabcolsep}{8pt}
\renewcommand{\arraystretch}{1.15}
\adjustbox{max width=\linewidth}{
\begin{tabular}{l c c}
\toprule
\textbf{Method} & \textbf{Macro F1}$\uparrow$ & \textbf{Macro FPR}$\downarrow$ \\
\midrule
\unilid (base)              & \textbf{\corrrev{0.960}} & \textbf{\corrrev{1.863e-4}} \\
\fasttext    & \corrrev{0.953} & 2.331e-4 \\
\hdashline
\unilid-Mistral-Nemo      & \corrrev{0.959} & \corrrev{1.894e-4} \\
\unilid-Mistral      & \corrrev{0.920} & \corrrev{3.380e-4} \\
\unilid-LLaMA3.2     & \corrrev{0.954} & \corrrev{2.081e-4}\\
\unilid-LLaMA2       & \corrrev{0.910}& \corrrev{3.733e-4} \\
\unilid-DeepSeek3.2       & \corrrev{0.955} & \corrrev{2.048e-4} \\
\unilid-Qwen3       & \corrrev{0.948} & \corrrev{2.341e-4} \\
\bottomrule
\end{tabular}}
\caption{Macro-averaged F1 and FPR of different LID systems on WiLI; \camrev{the rows below the dashed line} are variants of \unilid trained using the base tokenizers from open-source LLMs.}
\label{tab:unilid_llm_comparison}
\end{table}

\begin{table}[!htbp]
\centering
\small
\setlength{\tabcolsep}{8pt}
\renewcommand{\arraystretch}{1.15}
\caption{Comparison of Viterbi decoding versus exact marginalization (forward algorithm) at inference time on GlotLID-C. \corrrev{Inference under marginalization takes approximately $1.7\times$ the time of Viterbi decoding, and macro F1 is 0.002 higher.}}
\adjustbox{max width=\linewidth}{
\begin{tabular}{l c c}
\toprule
\textbf{Decoding} & \textbf{Accuracy}$\uparrow$ & \textbf{Macro F1}$\uparrow$ \\
\midrule
\unilid (Viterbi)         & 0.961 & \corrrev{0.933} \\
\unilid (Marginalization) & \corrrev{0.961} & \corrrev{\textbf{0.935}} \\
\bottomrule
\end{tabular}}
\label{tab:viterbi_vs_marginal}
\end{table}

\begin{table*}[!htbp]
\centering
\small
\setlength{\tabcolsep}{4pt}
\renewcommand{\arraystretch}{1.15}
\caption{\fasttext per-dialect F1 on DSL-ML 2024 across different training configurations. Macro is the unweighted mean F1 across all 14 dialects. All three BCMS dialects (HR, BS, ME) remain at 0.000 F1 regardless of training budget. \corrrev{Bold marks the best value in a column.}}
\adjustbox{max width=\linewidth}{
\begin{tabular}{l r r r r r r r r r r r r r r r}
\toprule
\textbf{Setting} & \textbf{FR-BE} & \textbf{FR-FR} & \textbf{FR-CA} & \textbf{FR-CH} & \textbf{ES-ES} & \textbf{ES-AR} & \textbf{PT-BR} & \textbf{PT-PT} & \textbf{EN-US} & \textbf{EN-GB} & \textbf{SR} & \textbf{HR} & \textbf{BS} & \textbf{ME} & \textbf{Macro} \\
\midrule
1 Epoch              & 0.6906 & 0.2325 & 0.0000 & 0.2989 & 0.0000 & 0.0000 & 0.0000 & 0.3936 & 0.0000 & 0.0000 & 0.0000 & 0.0000 & 0.0000 & 0.0000 & 0.1154 \\
5 Epochs             & 0.6894 & 0.3766 & 0.2786 & 0.3997 & 0.8133 & 0.0240 & 0.3812 & 0.5323 & 0.7410 & 0.0000 & 0.0000 & 0.0000 & 0.0000 & 0.0000 & 0.3026 \\
10 Epochs            & 0.6832 & 0.4180 & 0.7779 & 0.4676 & 0.8580 & 0.0000 & 0.4995 & 0.5793 & 0.7657 & 0.0000 & 0.0000 & 0.0000 & 0.0000 & 0.0000 & 0.3607 \\
10 Epochs, $MC$=10     & 0.6819 & 0.4163 & 0.8055 & \corrrev{\textbf{0.4956}} & 0.8292 & 0.0000 & 0.0243 & 0.5620 & 0.7725 & 0.0000 & 0.0000 & 0.0000 & 0.0000 & 0.0000 & 0.3277 \\
100 Epochs, $MC$=10    & 0.6908 & \textbf{0.4434} & \textbf{0.8139} & 0.4457 & \textbf{0.8674} & 0.5426 & 0.6088 & \textbf{0.6208} & 0.8065 & \textbf{0.5949} & 0.8293 & 0.0000 & 0.0000 & 0.0000 & 0.5189 \\
100 Epochs, $MC$=1000 & \textbf{.697} & .434 & .811 & .444 & .833 & \corrrev{\textbf{.686}} & \corrrev{\textbf{.729}} & .609 & \corrrev{\textbf{.815}} & .561 & \corrrev{\textbf{.833}} & .000 & .000 & .000 & \corrrev{\textbf{.532}} \\
\bottomrule
\end{tabular}}
\label{tab:fasttext_epoch_sweep}
\end{table*}

\subsection{Transfer to the \cld Label Subsets}
\label{app:cld3_refit}

\corrrev{The \cld-subset columns of \cref{tab:lid_main} compare systems each
fitted to the subset's languages. A second question is what the full
1{,}940-language systems score on the same slices, with predictions left
unrestricted so that an error into any of the 1{,}940 labels counts against the
gold language. \Cref{tab:cld3_refit} reports that transfer for \unilid and for its
calibrated form, beside the subset-fitted \unilid of the main table: transfer
costs 0.002 macro F1 on GlotLID-C, 0.009 on UDHR and 0.006 on FLORES-200. The
corrections applied to \cref{tab:lid_main}'s subset rows change each benchmark's macro F1 in the same
direction as they change that of the full system on the same slices: up on GlotLID-C,
down on UDHR, and on FLORES-200 by less than the table's printed precision. Most of that GlotLID-C gain is one language: Corsican
rises from F1 0.439 to 0.947, which is 0.0061 of the column's 0.0070.}

\begin{table}[htbp]
\centering
\small
\setlength{\tabcolsep}{5pt}
\renewcommand{\arraystretch}{1.15}
\adjustbox{max width=\linewidth}{
\begin{tabular}{l cc cc cc}
\toprule
& \multicolumn{2}{c}{\textbf{GlotLID-C} \small(83 langs)}
& \multicolumn{2}{c}{\textbf{UDHR} \small(80 langs)}
& \multicolumn{2}{c}{\textbf{FLORES-200} \small(77 langs)} \\
\cmidrule(lr){2-3} \cmidrule(lr){4-5} \cmidrule(lr){6-7}
\textbf{System} & F1$\uparrow$ & FPR$\downarrow$ & F1$\uparrow$ & FPR$\downarrow$ & F1$\uparrow$ & FPR$\downarrow$ \\
\midrule
\unilid, transferred                        & .972 & 9.99e-5 & .986 & 1.64e-5 & .991 & 3.50e-5 \\
\unilid calibrated, transferred             & .975 & 1.24e-4 & .985 & 2.11e-5 & .992 & 3.83e-5 \\
\midrule
\unilid, fitted to the subset               & .974 & 1.52e-4 & .995 & 5.73e-5 & .997 & 3.93e-5 \\
\bottomrule
\end{tabular}}
\caption{\unilid on the \cld label subsets, when using a model trained on all GlotLID-C languages rather than only the \cld-subset. The \cld-subset columns of \cref{tab:lid_main} instead compare systems each fitted to the subset's languages. Predictions here are left unrestricted for \unilid, so that an error into any of the 1{,}940 labels counts against
the gold language. 
The calibrated system appears in both forms: transferred
here, and fitted to each subset in \cref{tab:lid_main}. The last row is the
subset-fitted \unilid of \cref{tab:lid_main}, repeated for comparison. }
\label{tab:cld3_refit}
\end{table}

\subsection{Viterbi vs. Marginalization at Inference}
\label{app:viterbi_vs_marginal}

\unilid uses Viterbi decoding at inference (\cref{sec:inference}), scoring each language by its most likely segmentation. An alternative is to compute the exact marginal likelihood $p(s \mid \ell)$ by summing over all segmentations using the forward algorithm. \Cref{tab:viterbi_vs_marginal} compares the two strategies on GlotLID-C. \camrev{The marginalized variant scores 0.002 higher macro F1 (\corrrev{0.935 against 0.933}) \corrrev{and the same accuracy to three decimals, at approximately $1.7\times$ the inference cost}; we did not estimate the uncertainty of this comparison. Given the small difference, we retain Viterbi as the default inference method.}

\begin{table}[t]
\centering
\small
\caption{Accuracy as a function of input sequence length for \unilid and \fasttext on WiLI. All samples in WiLI are $>$ 100 chars long.}
\label{tab:length_accuracy}
\setlength{\tabcolsep}{6pt}
\renewcommand{\arraystretch}{1.15}
\adjustbox{max width=0.8\linewidth}{
\begin{tabular}{l r c c}
\toprule
\textbf{Length (chars)} & \textbf{Samples} & 
\begin{tabular}[c]{@{}c@{}}\textbf{\unilid}\\ Acc. (\%)$\uparrow$\end{tabular}
  & \begin{tabular}[c]{@{}c@{}}\textbf{\fasttext}\\ Acc. (\%)$\uparrow$\end{tabular}\\
\midrule101--150   & 7,845   & \textbf{\corrrev{93.04}} & 90.73 \\
151--200   & 26,652  & \textbf{\corrrev{94.11}} & 92.56 \\
201--300   & 31,449  & \textbf{\corrrev{95.83}} & 94.58 \\
301--500   & 29,494  & \textbf{\corrrev{96.79}} & 96.03 \\
501--1000  & 18,142  & \textbf{\corrrev{96.60}} & 96.25 \\
1000+      & 3,918   & \textbf{\corrrev{96.61}} & 96.30 \\
\midrule
\textbf{Overall} & 117,500 & \textbf{\corrrev{95.64}} & 94.54 \\
\bottomrule
\end{tabular}}
\end{table}

\subsection{Robustness to Orthographic Noise}
\label{app:noise_robustness}
\begin{table*}[!htbp]
\centering
\small
\setlength{\tabcolsep}{6pt}
\renewcommand{\arraystretch}{1.15}
\caption{Performance under stochastic character-level perturbation on WiLI (117,500 samples, 235 languages). Each non-whitespace character is replaced with probability $p$. \unilid maintains a small edge at low noise; \fasttext degrades more gracefully at higher noise rates.}
\adjustbox{max width=\linewidth}{
\begin{tabular}{c ccc ccc}
\toprule
& \multicolumn{3}{c}{UniLID} & \multicolumn{3}{c}{fastText} \\
\cmidrule(lr){2-4} \cmidrule(lr){5-7}
Noise $p$ & Acc & F1 & FPR & Acc & F1 & FPR \\
\midrule
0\%  & $\mathbf{0.9564}$ & $\mathbf{0.9601}$ & $\mathbf{1.863\times10^{-4}}$ & $0.9527$ & $0.9533$ & $2.023\times10^{-4}$ \\
5\%  & $\mathbf{0.9520}\pm0.0002$ & $\mathbf{0.9553}\pm0.0002$ & $\mathbf{2.051\times10^{-4}}$ & $0.9474\pm0.0003$ & $0.9479\pm0.0003$ & $2.249\times10^{-4}$ \\
10\% & $\mathbf{0.9439}\pm0.0003$ & $\mathbf{0.9468}\pm0.0003$ & $\mathbf{2.398\times10^{-4}}$ & $0.9401\pm0.0007$ & $0.9406\pm0.0007$ & $2.559\times10^{-4}$ \\
25\% & $0.8562\pm0.0003$ & $0.8586\pm0.0001$ & $6.147\times10^{-4}$ & $\mathbf{0.8925}\pm0.0005$ & $\mathbf{0.8922}\pm0.0005$ & $\mathbf{4.592\times10^{-4}}$ \\
50\% & $0.4749\pm0.0009$ & $0.4750\pm0.0010$ & $2.244\times10^{-3}$ & $\mathbf{0.6256}\pm0.0005$ & $\mathbf{0.6272}\pm0.0004$ & $\mathbf{1.600\times10^{-3}}$ \\
\bottomrule
\end{tabular}}
\label{tab:noise_robustness}
\end{table*}
 To evaluate robustness under realistic input corruption, we apply stochastic character-level perturbations to the WiLI test set: with probability $p$, each non-whitespace character is independently replaced with another character drawn uniformly from the inventory of characters observed in the WiLI training set. We evaluate at $p \in \{0\%, 5\%, 10\%, 25\%, 50\%\}$. Example inputs at $p=5\%$ and $p=10\%$ are shown below; \cref{tab:noise_robustness} reports accuracy, macro F1, and macro FPR for \unilid and \fasttext on the perturbed test sets. \corrrev{The \fasttext numbers in the $p=0\%$ row come from a separate evaluation run and therefore differ slightly from the \fasttext WiLI numbers reported elsewhere in the paper.}

\paragraph{Example perturbations.}
\begin{description}
    \item[Original] \emph{Anton (or Antonius) Maria Schyrleus (also Schyrl, Schyrle) of Rheita (1604--1660) was an astronomer and optician. He developed several inverting and erecting eyepieces\ldots}
    \item[$p=5\%$] \emph{Anton (or Antonius) Maria Schyrlezs $\ldots$ Anton\'in $\ldots$ astronmmer and optijian\ldots}
    \item[$p=10\%$] \emph{Anton (or Aston,us) MDria SchyrKeus $\ldots$ of iheith\ldots}
\end{description}

At low noise ($p < 10\%$), \unilid maintains a small edge in F1 and accuracy. At moderate-to-high noise ($p \ge 25\%$), \fasttext degrades more gracefully than \unilid: the character n-gram representations underlying \fasttext appear to absorb localized character corruptions more robustly than \unilid's segmentation-based scoring, where corrupted characters can fragment otherwise-high-probability tokens. This points to a possible avenue for future work --- explicit noise-aware token scoring or character-level smoothing within the \unilid framework.
\section{\fasttext Hyperparameter Sensitivity on Dialect Identification}
\label{app:fasttext_epochs}

To validate the \fasttext configuration used in our dialect experiments (\cref{sec:results}), we ran a hyperparameter sweep over the number of training epochs and the minimum word-count threshold ($MC$) on the DSL-ML 2024 train/dev splits. \Cref{tab:fasttext_epoch_sweep} reports per-dialect F1 along with the macro-averaged F1 across all 14 labels. \corrrev{Macro F1 rises with the number of epochs at the default minimum word count, from 0.1154 at 1 epoch to 0.3026 at 5 and 0.3607 at 10, and the best configuration in the sweep is 100 epochs with \texttt{min-count}=1000, at 0.532. The one step that does not follow this pattern is \texttt{min-count}=10 at 10 epochs, which scores 0.3277, below the 0.3607 of 10 epochs at the default minimum word count.} We use \corrrev{the 100-epoch, \texttt{min-count}=1000} configuration as our \fasttext baseline throughout the paper. Crucially, the three South Slavic dialects (HR, BS, ME) remain at 0.000 F1 across \emph{every} \fasttext configuration in the sweep, indicating that \fasttext fails on these closely related dialects regardless of training budget.

\section{Inference Latency and Training Time}
\label{app:efficiency}

For reference, \fasttext and \cld share a similar input-processing cost linear in \corrrev{the input string length $T$}: \fasttext extracts character n-grams over a sliding window (\corrrev{$\mathcal{O}(T \cdot k)$} for window size $k$).
It then averages n-gram embeddings (\corrrev{$\mathcal{O}(T \cdot d))$} and scores all languages with a single matrix multiplication ($\mathcal{O}(d \cdot|\languages|)$); \cld's scoring is asymptotically equivalent. The structural difference is that the language-scoring step in \fasttext and \cld is independent of input length, whereas \corrrev{\unilid's scales with $T$}. In practice this is offset by the fact that the segmentation lattice is constructed once and reused across languages.

\Cref{tab:latency_glotlid,tab:latency_wili} \camrev{report} end-to-end inference latency and throughput for \unilid, \fasttext, and \cld measured on the same hardware. On GlotLID-C ($|\languages|=1940$), \unilid is approximately \camrev{$1.85\times$} slower than \fasttext per sample; on WiLI ($|\Lambda|=235$), \camrev{the difference narrows to $1.4\times$} and \unilid is over $2.7\times$ faster than \cld. \Cref{tab:training_time} compares wall-clock training time for the GlotLID-C training set.

\begin{table}[!htbp]
\centering
\small
\setlength{\tabcolsep}{8pt}
\renewcommand{\arraystretch}{1.15}
\caption{Inference latency on GlotLID-C ($|\Lambda|=1940$), evaluated over 45,627,279 samples.}
\adjustbox{max width=\linewidth}{
\begin{tabular}{l r r}
\toprule
\textbf{Method} & \textbf{Latency (ms/sample)}$\downarrow$ & \textbf{Throughput (samples/s)}$\uparrow$ \\
\midrule
\fasttext    & \textbf{0.166} & \textbf{6019} \\
\cld         & 0.187 & 5356 \\
\unilid      & 0.307 & 3253 \\
\bottomrule
\end{tabular}}
\label{tab:latency_glotlid}
\end{table}

\begin{table}[!htbp]
\centering
\small
\setlength{\tabcolsep}{8pt}
\renewcommand{\arraystretch}{1.15}
\caption{Inference latency on WiLI ($|\Lambda|=235$), evaluated over 117,500 samples.}
\adjustbox{max width=\linewidth}{
\begin{tabular}{l r r}
\toprule
\textbf{Method} & \textbf{Latency (ms/sample)}$\downarrow$ & \textbf{Throughput (samples/s)}$\uparrow$ \\
\midrule
\fasttext    & \textbf{0.113} & \textbf{8889} \\
\unilid      & 0.155 & 6435 \\
\cld         & 0.427 & 2340 \\
\bottomrule
\end{tabular}}
\label{tab:latency_wili}
\end{table}

\begin{table}[!htbp]
\centering
\small
\setlength{\tabcolsep}{8pt}
\renewcommand{\arraystretch}{1.15}
\caption{Wall-clock training time on the GlotLID-C training set. \unilid splits the training cost across base tokenizer learning (1k samples per language) and per-language EM re-estimation; \fasttext is trained on the full training set for 100 epochs.}
\adjustbox{max width=\linewidth}{
\begin{tabular}{l r}
\toprule
\textbf{Method} & \textbf{Total Time (s)}$\downarrow$ \\
\midrule
\unilid              & \textbf{17{,}776} \\
\fasttext (100 ep.)  & $\sim$163{,}000 \\
\bottomrule
\end{tabular}}
\label{tab:training_time}
\end{table}

\section{Performance Breakdown}\label{app:perf-breakdown} In order to understand its strengths and weaknesses, we analyze \unilid's performance along three axes: writing script, training-data quantity, and segmentation length under predicted vs. true language.
\subsection{Script}
\cref{tab:script-breakdown} partitions GlotLID-C test performance by Unicode script. UniLID performs almost identically to \fasttext on Latin-script languages (1,700 of the 1,940 labels; F1 \corrrev{0.944} vs. 0.946), but underperforms on every non-Latin script. The gaps are largest on Greek ($\Delta$ F1 = \corrrev{-0.250}), Hebrew (\corrrev{-0.229}), Devanagari (-0.121), and Bengali (\corrrev{-0.106}). \camrev{We attribute this to the base vocabulary ($\vocab$) of the tokenizer: $\vocab$ is learned by \unilm on 1{,}000 samples per language of the GlotLID-C training corpus, in which 1{,}700 of the 1{,}940 languages are Latin-script, so most of the vocabulary's training text is Latin-script.} \corrrev{A vocabulary learned mostly from Latin-script text contains fewer multi-character tokens for other scripts. Per-language estimation can only reweight the tokens that exist in $\vocab$, so strings in those scripts are segmented into shorter tokens.}
Larger vocabularies help modestly (\cref{app:additional_results}) but do not close the gap. The most direct solution would be  to train $\vocab$ on a script-balanced or script-upsampled corpus so that non-Latin substrings receive proportional coverage in $\vocab$. We leave this exploration to future work. 
\begin{table}[t]
\centering
\small
\begin{tabular}{@{}lrrrr@{}}
\toprule
Script & \# Langs & UniLID & fastText & $\Delta$ \\
\midrule
Latn  & 1{,}700 & \corrrev{0.944} & 0.946 & \corrrev{$-0.002$} \\
Cyrl  &      70 & \corrrev{0.880} & 0.970 & \corrrev{$-0.090$} \\
Arab  &      38 & \corrrev{0.693} & 0.747 & \corrrev{$-0.054$} \\
Deva  &      32 & 0.811 & 0.932 & $-0.121$ \\
Beng  &       6 & \corrrev{0.879} & 0.985 & \corrrev{$-0.106$} \\
Grek  &       4 & \corrrev{0.675} & 0.925 & \corrrev{$-0.250$} \\
Hebr  &       4 & \corrrev{0.738} & 0.967 & \corrrev{$-0.229$} \\
Armn  &       2 & \corrrev{0.972} & 0.986 & \corrrev{$-0.014$} \\
Other &      82 & 0.937 & 0.973 & $-0.036$ \\
\bottomrule
\end{tabular}
\caption{Macro-F1 on the GlotLID-C test set stratified by Unicode script (ISO 15924), in the within-stratum view: test examples are restricted to true labels inside each script group, so false positives arriving from other scripts are excluded. The two single-language scripts (Korean, Japanese) are excluded from the Other row. $\Delta = \text{F1}(\text{UniLID}) - \text{F1}(\text{fastText})$. UniLID matches fastText on Latin-script languages but underperforms on every non-Latin script, with the largest gaps on Greek, Hebrew, and Devanagari. We attribute this to the training distribution of the base vocabulary $V$, which is dominated by Latin-script text in GlotLID-C.}
\label{tab:script-breakdown}
\end{table}

\subsection{Resource Tier}
\cref{tab:resource-tier} stratifies results by per-language training-sample count. \camrev{\unilid matches or exceeds \fasttext for languages with 500$+$ training samples.} The differences arise in the very low resource settings: \unilid lags \fasttext on languages with fewer than 500 training samples, \corrrev{by 0.058 macro F1 in the within-stratum view of this table (0.857 against 0.915) and by 0.154 in the global view of \cref{tab:calibrated_views} (0.596 against 0.750)}. \corrrev{Many of the low-resource languages in GlotLID-C are written in non-Latin scripts, so this tier confounds training-sample count with script: many of the languages below 500 training samples are also languages for which $\vocab$ holds few multi-character tokens (\cref{tab:script-breakdown}), and the tier result therefore cannot be attributed to sample count alone. Our WiLI subsampling experiments (\cref{sec:results}) do not separate the two either, because they reduce the training data of all 235 WiLI languages at once, across scripts.}

\begin{table*}[t]
\centering
\small
\begin{tabular}{@{}lrrcccc@{}}
\toprule
 & & & \multicolumn{2}{c}{UniLID} & \multicolumn{2}{c}{fastText} \\
\cmidrule(lr){4-5} \cmidrule(lr){6-7}
Train samples / lang & \# Langs & $N_{\text{test}}$ & F1 & FPR & F1 & FPR \\
\midrule
$<\!500$         &  56 & \num{2513}     & \corrrev{0.857} & \corrrev{\num{6.5e-5}} & 0.915 & \num{1.15e-4} \\
$500$\,--\,$1$k    &  40 & \num{5222}     & \corrrev{0.973} & \corrrev{\num{1.0e-5}} & 0.964 & \num{1.9e-5}  \\
$1$k\,--\,$12$k    & 458 & \num{552346}   & 0.990 & \num{8.0e-6} & 0.979 & \num{8.0e-6}  \\
$12$k\,--\,$18$k   & 526 & \num{1151363}  & 0.997 & \num{2.0e-6} & 0.986 & \num{1.0e-5}  \\
$18$k\,--\,$35$k   & 398 & \num{1125105}  & 0.992 & \num{7.0e-6} & 0.981 & \num{1.6e-5}  \\
$35$k$+$           & 462 & \num{42540730} & 0.958 & \corrrev{\num{5.4e-5}} & 0.942 & \num{9.1e-5}  \\
\bottomrule
\end{tabular}
\caption{Performance on the GlotLID-C test set stratified by per-language training-sample count, in the within-stratum view: test examples are restricted to true labels inside each tier, so false positives arriving from other tiers are excluded. \Cref{tab:calibrated_views} repeats this stratification under both metric views. \# Langs is the number of languages in the bucket; $N_{\text{test}}$ is the total number of test instances \camrev{on the 45{,}377{,}279-line scored pool (\cref{app:protocol})}. UniLID matches or exceeds fastText on every bucket except $<\!500$, where the base-vocabulary script bias documented in \cref{tab:script-breakdown} compounds with limited per-language data.}
\label{tab:resource-tier}
\end{table*}
 
\subsection{Length Bias}
\camrev{We lastly look at potential length biases.} The unigram likelihood lacks an explicit length prior, so a language whose $\unigramdistlang$ puts more mass on longer tokens accumulates fewer multiplicative factors than one favoring shorter tokens. 
We checked whether this manifests as a systematic bias in errors and find that on misclassifications, the Viterbi segmentation under the predicted language is on average \corrrev{0.05} tokens shorter than under the true language, \corrrev{with that difference growing only for inputs above 150 characters}. Length-normalizing the per-language log-likelihood by segmentation length led to worse \camrev{performance} in preliminary experiments; we report results in \cref{tab:lenbias-delta,tab:lenbias-norm}.

\begin{table*}[htbp]
\centering
\small
\begin{tabular}{lrrrrrr}
\toprule
\textbf{Length (chars)} & \textbf{N} & \textbf{Mean $\Delta$} & \textbf{Median $\Delta$} & \textbf{\% fewer} & \textbf{\% same} & \textbf{\% more} \\
\midrule
All misclassified & \corrrev{9,906} & \corrrev{$-0.05$} & $0.00$ & \corrrev{25.29} & \corrrev{52.12} & \corrrev{22.59} \\
\midrule
$<$30    &   \corrrev{2,814} & \corrrev{$-0.05$} & $0.00$ & \corrrev{19.76} & \corrrev{65.28} &  \corrrev{14.96} \\
30--75   &   \corrrev{4,316} & \corrrev{$-0.01$} & $0.00$ & \corrrev{23.96} & \corrrev{53.59} & \corrrev{22.45} \\
75--150  &   \corrrev{2,206} & \corrrev{$-0.04$} & $0.00$ & \corrrev{31.14} & \corrrev{39.30} & \corrrev{29.56} \\
150--300 &   \corrrev{532} & \corrrev{$-0.23$} & $0.00$ & \corrrev{39.66} & \corrrev{26.50} & \corrrev{33.83} \\
300+     &     \corrrev{38} & \corrrev{$-2.13$} & \corrrev{$0.00$} & \corrrev{44.74} & \corrrev{13.16} & \corrrev{42.11} \\
\bottomrule
\end{tabular}
\caption{%
  Token-count length bias on \unilid misclassifications, by input length.
  For each misclassified sample, $\Delta = n_{\mathrm{pred}} - n_{\mathrm{true}}$ is the
  difference in Viterbi segmentation length (in tokens) between the predicted and the true
  language; negative values mean the predicted language uses fewer tokens. \corrrev{Computed over the
  $9{,}906$ misclassifications in the $250{,}000$-line golden subset, the test half of the
  seed-42 $500{,}000$-line draw and the same subset as \cref{tab:lenbias-norm}; the N column is
  therefore not comparable with a count over the full test set.} The mean
  $\Delta$ is negative in every row and \corrrev{grows in magnitude only for inputs above $150$
  characters}, while the median is zero and \corrrev{$52\%$} of errors leave the token count unchanged: a small but systematic bias toward
  fewer-token languages, driven by a minority of samples. \emph{\% fewer/same/more} give the
  share of misclassifications where the predicted language uses fewer, equal, or more tokens
  than the true language. 
}
\label{tab:lenbias-delta}
\end{table*}

\begin{table}[htbp]
\centering
\small
\adjustbox{max width=\columnwidth}{
\begin{tabular}{lrrrr}
\toprule
\textbf{Length (chars)} & \textbf{N} & \textbf{Original} & \textbf{Raw rescore} & \textbf{Normalized} \\
\midrule
$<$30    &  \corrrev{13,708} & \corrrev{0.795} & \corrrev{0.795} & \corrrev{0.494} \\
30--75   & \corrrev{88,503} & \corrrev{0.951} & \corrrev{0.951} & \corrrev{0.776} \\
75--150  & \corrrev{97,861} & \corrrev{0.977} & \corrrev{0.977} & \corrrev{0.883} \\
150--300 &  \corrrev{43,566} & \corrrev{0.988} & \corrrev{0.988} & \corrrev{0.946} \\
300+     &  \corrrev{6,362} & \corrrev{0.994} & \corrrev{0.994} & \corrrev{0.986} \\
\midrule
Overall  & \corrrev{250,000} & \corrrev{0.960} & \corrrev{0.960} & \corrrev{0.837} \\
\bottomrule
\end{tabular}}
\caption{%
  Effect of length-normalizing the per-language log-likelihood on \unilid accuracy, by input
  length. Scores are normalized by dividing the summed token log-probabilities by the
  segmentation length ($\mathrm{score}/n_{\mathrm{tokens}}^{\alpha}$ with $\alpha=1$).
  \emph{Raw rescore} re-runs the unnormalized scorer ($\alpha=0$) through the same code path
  and reproduces the original predictions exactly ($100\%$ agreement), validating the
  implementation. \corrrev{Evaluated on the $250k$-line test half of a $500k$-line uniform sample.} Full normalization lowers overall accuracy from $0.960$ to \corrrev{$0.837$}, with the
  largest degradation on short inputs ($<$30 chars: \corrrev{$0.795 \to 0.494$}) and almost none on long
  inputs: the opposite of where the token-count bias in \cref{tab:lenbias-delta} is
  largest. 
}
\label{tab:lenbias-norm}
\end{table}
 \section{Development protocol and additional measurements for calibrated \unilid}\label{app:protocol}

\paragraph{Full mechanism specification.} Calibrated \unilid
(\cref{sec:calibration}) re-examines a prediction when (i) the predicted
language has fewer than 18{,}000 training samples, or is one of the four
high-entropy languages defined next, and (ii) \corrrev{the line's
log-probability margin is smaller than the predicted language's threshold
$\delta_\lang$}. A language belongs to the high-entropy group when the entropy
of its estimated distribution $\unigramdistlang$, standardized among the
languages sharing its script (\corrrev{the entropy minus the median entropy of
the languages in that script, divided by their median absolute deviation}), exceeds $1.5$ AND the language received more
than twice as many misclassified \corrrev{selection-half} samples as it has \corrrev{selection-half}
samples of its own, or when the standardized entropy alone exceeds $5$;
restricted to languages with at least 18{,}000 training samples (so the two
re-examined groups are disjoint), this selects Scots, Banjar, Aragonese, and
West Flemish. \camrev{The \corrrev{log-probability} margin of a line is the top score minus the
runner-up score, computed under the distributions with the unseen-token
constant applied.} The percentile defining \corrrev{$\delta_\lang$} is
$q_\lang = 5\,(1 - N_\lang/18{,}000)$ for a language with $N_\lang$ training
samples, and a fixed 5th percentile for the high-entropy group\camrev{; each
threshold is estimated from the margins on up to 2{,}000 of the language's
own training lines on which that language is the top-scoring candidate
(fixed seed), and the 26 of the 1{,}080 languages below 18{,}000 samples
with fewer than 200 such lines receive no threshold and are never
re-examined. Because $q_\lang$ declines linearly to zero at 18{,}000
samples, the re-examined fraction of a language's own predictions shrinks
to zero as its training count approaches the boundary.} A
re-examined prediction is reassigned to the highest-ranked of the candidates
ranked 2 to 5 that has at least 100{,}000 training samples and a score
within 21 natural-log units of the top candidate; if no candidate qualifies,
the prediction is unchanged. Under the unseen-token constant alone, with no
re-examination, held-out macro F1 rises from \corrrev{0.916} to 0.930.

\paragraph{Provenance of the constants.} The unseen-token constant
\corrrev{$c = -17$ was chosen from a sweep over $\{-15, -17, -19, -21\}$} on \camrev{the
\corrrev{selection half} described in the development protocol below;
the proximity bound $21$ from a grid search over values from $0.5$ to $100$
on the development part; overall macro F1 on that part varies by less than
$0.0003$ across bounds from roughly $15$ to $35$, so $21$ is a
representative value of that range rather than a finely tuned one;} the
$100{,}000$-sample \camrev{requirement} from an analysis of where reassignments send their
false positives\camrev{; this requirement coincides with the per-language
training-data cap of \cref{app:datasets}, so it admits exactly the 282
languages whose training data reached the cap. The remaining constants (the
$18{,}000$-sample boundary, the entropy z-score thresholds of $1.5$ and
$5$, the absorption factor $2$, the fixed 5th percentile, the form of
$q_\lang$, and the rank cutoff of $5$) were fixed during development; no
search over alternatives was recorded for them}. The
\corrrev{misclassified-sample count} in the
high-entropy-group criterion is computed on \camrev{the \corrrev{selection
half}}. All
constants are frozen before evaluation on the held-out reporting subset,
which was excluded from every one of these selections except the
$100{,}000$-sample \camrev{requirement}, fixed from a false-positive profile measured on the
whole \camrev{45.0M-line pool remaining after the two draws were removed,}
before this split existed.

\paragraph{Development protocol.} \camrev{The GlotLID-C test data contains
45{,}627{,}279 lines. \corrrev{A $500{,}000$-line sample was drawn uniformly
from them with a fixed seed and split into two halves of $250{,}000$ lines. We
call the first the \defn{selection half}: it was set aside early in development
and used only to select constants, and it is excluded from the GlotLID-C
numbers we report for \unilid and calibrated \unilid, leaving
$45{,}377{,}279$ scored lines. We call the second the \defn{diagnostic half};
it stays inside the scored pool and is the subset on which
\cref{tab:lenbias-delta,tab:lenbias-norm} are computed.}} The
constants were selected on data disjoint from
the subset used to report their effect. From the 45{,}377{,}279 scored lines
of the GlotLID-C test pool, we removed two balanced draws (188{,}061 and
185{,}204 lines, at most 100 lines per language, seeds 101 and 201) and split
the remaining 45{,}004{,}014 lines by a seeded 40/60 partition (seed 301)
into a development part (18{,}001{,}573 lines) and a held-out part
(27{,}002{,}441 lines). The unseen-token constant was selected on \camrev{the
\corrrev{selection half}}; the proximity bound on the
development part; the re-examination thresholds on each language's own
training samples. The held-out part was used only to confirm each candidate
configuration once, and it is the subset on which the paired bootstrap
intervals in this appendix are computed. \Cref{tab:calibration_provenance}
lists each selected component with its selection data.
\cref{tab:calibrated_heldout} repeats the comparison on the held-out part.

\paragraph{Provenance of the main results table.} \corrrev{The
GlotLID-C cells of the \unilid and calibrated \unilid rows of
\cref{tab:lid_main} are macro-averaged
over the 1{,}940 labels on the $45{,}377{,}279$ scored lines defined above. The
GlotLID-C cells of the remaining rows are computed on all $45{,}627{,}279$
test lines; \fasttext macro F1 on the smaller pool differs from the value
printed for the full file by $3.4\times10^{-4}$. The calibrated row's
full-pool GlotLID-C cells include the development part, on which two of the
calibration constants were selected, which is why \cref{tab:calibrated_heldout}
repeats the comparison on the held-out part alone. In the \cld-subset columns
the \unilid row comes from models whose base vocabulary and per-language
distributions are trained only on that subset's languages, by the procedure of
\cref{sec:exp_setup}, and \fasttext is likewise trained on the subset languages
alone. The \cld row is the pretrained model of \cref{sec:exp_setup}, scored with
its predictions restricted to the subset's labels, so \unilid, \fasttext and
\cld predict over the same label set in these columns. The calibrated row's \cld-subset cells apply the procedure of
\cref{sec:calibration} to each subset model with every constant fixed on the
full model and carried unchanged; no hyperparameter was swept for the subset,
and only the language groups and the per-language thresholds are recomputed,
which the procedure defines as functions of the model it is applied to. The
high-entropy group is empty on all three subsets, so re-examination there
applies to the low-resource group alone. The three \unilid-X rows use fixed
pretrained vocabularies, which cannot be fitted to a subset, so their
\cld-subset cells are not comparable with the subset-fitted \unilid row.}

\begin{table}[htbp]
\centering
\small
\adjustbox{max width=\linewidth}{
\begin{tabular}{lll}
\toprule
\textbf{Component} & \textbf{Selection data} & \textbf{In scored pool?} \\
\midrule
unseen-token constant \corrrev{$c=-17$} & \camrev{excluded 250k validation sample} & no \\
high-entropy-group membership & \camrev{same excluded 250k sample} & no \\
thresholds, $q_\lang$ & each language's own training data & no \\
$100{,}000$-sample \camrev{requirement} & false-positive profile, full pool & yes \\
proximity bound $21$ & development part (18.0M lines) & yes \\
\bottomrule
\end{tabular}}
\caption{Selected components of calibrated \unilid and the data each was
selected on. The held-out part is outside every row of this table except the
$100{,}000$-sample \camrev{requirement}, which was fixed from a false-positive profile measured
on the whole \camrev{45.0M-line pool remaining after the two draws were
removed,} before the development/held-out split existed.}
\label{tab:calibration_provenance}
\end{table}

\begin{table}[htbp]
\centering
\small
\begin{tabular}{lrr}
\toprule
\textbf{Method} & \textbf{Macro F1} & \textbf{Ma-FPR (x1e5)} \\
\midrule
\unilid & \corrrev{0.916} & \corrrev{2.03} \\
calibrated \unilid & \corrrev{0.949} & \corrrev{1.78} \\
\fasttext & 0.933 & 2.72 \\
\bottomrule
\end{tabular}
\quad
\begin{tabular}{lrr}
\toprule
\textbf{Comparator} & \textbf{Mean diff} & \textbf{95\% CI} \\
\midrule
\unilid & \corrrev{$+0.034$} & \corrrev{$[+0.029, +0.038]$} \\
\fasttext & \corrrev{$+0.016$} & $[+0.011, +0.022]$ \\
\bottomrule
\end{tabular}
\caption{Left: macro F1 and macro FPR on the held-out part (27{,}002{,}441
lines), outside every calibration-constant selection except the
$100{,}000$-sample \camrev{requirement} (see the provenance table). Right: paired
bootstrap of calibrated \unilid minus each comparator on the held-out part
(10{,}000 resamples over the 1{,}940 languages, percentile interval). FPR
values are multiplied by $10^5$. \camrev{Macro F1 is not invariant to
subsampling, so the held-out levels sit below the full-pool levels of
\cref{tab:lid_main} for every system; the quantity of interest here is the
paired difference.}}
\label{tab:calibrated_heldout}
\end{table}

\begin{table*}[htbp]
\centering
\small
\begin{tabular}{lrrrrrrr}
\toprule
& & \multicolumn{3}{c}{\textbf{Global view}} & \multicolumn{3}{c}{\textbf{Within-stratum view}} \\
\cmidrule(lr){3-5} \cmidrule(lr){6-8}
\textbf{Train samples} & \textbf{\# Langs} & \unilid & calib.\ \unilid & \fasttext & \unilid & calib.\ \unilid & \fasttext \\
\midrule
$<$500 & 56 & \corrrev{0.596} & \corrrev{0.781} & 0.750 & \corrrev{0.857} & \corrrev{0.820} & 0.915 \\
500--1k & 40 & \corrrev{0.676} & \corrrev{0.893} & 0.861 & \corrrev{0.973} & \corrrev{0.954} & 0.964 \\
1k--12k & 458 & \corrrev{0.894} & \corrrev{0.944} & 0.941 & 0.990 & \corrrev{0.986} & 0.979 \\
12k--18k & 526 & 0.979 & 0.985 & 0.971 & 0.997 & 0.997 & 0.986 \\
18k--35k & 398 & \corrrev{0.962} & 0.965 & 0.953 & 0.992 & 0.992 & 0.981 \\
35k+ & 462 & \corrrev{0.957} & \corrrev{0.956} & 0.941 & 0.958 & \corrrev{0.957} & 0.942 \\
\bottomrule
\end{tabular}
\caption{Mean per-language F1 by training-resource tier under both metric
views, on the 45{,}377{,}279 scored test lines. Global view: every false
positive counts against the predicted language. Within-stratum view: test
examples are restricted to true labels inside the tier, so false positives
arriving from other tiers are excluded (the view used by
\cref{tab:resource-tier}). The two views rank the methods oppositely in the
smallest tiers: calibrated \unilid removes false positives \camrev{\emph{into}} small
languages, which only the global view counts, and re-examination moves some
of the smallest languages' own low-margin lines away, which only the
within-stratum view penalizes.}
\label{tab:calibrated_views}
\end{table*}
 
\paragraph{\camrev{Measurements on other evaluation sets.}} On a balanced draw of the test
pool (draw seed 201, 185{,}204 lines, at most 100 lines per language), the
within-stratum overall macro F1 of calibrated \unilid is 0.978 against 0.981
for \unilid: a balanced draw removes most of the false-positive volume that
\camrev{\corrrev{calibrated \unilid} corrects}, so the gain does not appear there.
\Cref{tab:commonlid} reports an out-of-domain web-domain evaluation
(CommonLID): \corrrev{calibrated \unilid} raises macrolanguage-aware accuracy from \corrrev{0.848} to
\corrrev{0.862} while tag-level macro F1 \corrrev{falls} from \corrrev{0.722} to \corrrev{0.717}\camrev{:
\corrrev{calibrated \unilid} reduces predictions into languages outside the 109-tag set
(\corrrev{32{,}525} to \corrrev{25{,}994} lines), which raises accuracy, while the
re-examination also \corrrev{reassigns} some correct low-margin lines, which lowers
tag-level macro F1. Both measurements point to the setting in which the
method's gains should be expected: test data whose per-language line counts
follow the collection's natural imbalance, over a label set that includes
under-resourced languages.}

\begin{table}[htbp]
\centering
\small
\adjustbox{max width=\linewidth}{
\begin{tabular}{lrr}
\toprule
\textbf{Method} & \textbf{Accuracy} & \textbf{Tag-level macro F1} \\
\midrule
\unilid & \corrrev{0.848} & \corrrev{0.722} \\
\unilid, unseen-token constant only & \corrrev{0.851} & \corrrev{0.720} \\
calibrated \unilid & \corrrev{0.862} & \corrrev{0.717} \\
\bottomrule
\end{tabular}}
\caption{Out-of-domain evaluation on CommonLID (373{,}230 web-domain lines,
109 bare ISO 639-3 tags). A prediction is correct if its language code equals
the gold tag or is a member of the gold macrolanguage; accuracy is
macrolanguage-aware, and tag-level macro F1 averages over the 109 tags.
\corrrev{Calibrated \unilid} raises accuracy because predictions into languages outside the
109-tag set fall from \corrrev{32{,}525} to \corrrev{25{,}994} \camrev{lines}; tag-level macro F1 decreases
slightly because the re-examination also \corrrev{reassigns} some correct low-margin lines,
and the under-resourced languages \corrrev{the two corrections repair} are largely absent
from this tag set.}
\label{tab:commonlid}
\end{table}
 
\paragraph{Alternatives measured during development.} (i) \corrrev{Setting} each language's unseen-token
mass \corrrev{to} its own Good-Turing estimate \corrrev{(the estimate that assigns the
unseen tokens the total probability of the tokens seen exactly once)}, instead of the shared constant,
raises false positives into the under-1{,}000-sample languages from
22{,}404 to 79{,}113 \camrev{(counted over the 45.0M-line pool remaining
after the two draws were removed)}: each
language's own \corrrev{Good-Turing estimate} is \camrev{accurate} in isolation, but the cross-language
comparison needs the shared value. (ii) Applying the unseen-token constant
only to languages with fewer than 18{,}000 training samples (leaving the
larger languages' unseen-token values as trained) \corrrev{changes held-out
macro F1 by $+0.0002$ with 95\% interval $[-0.0003, +0.0006]$; the interval
contains zero, so this measurement does not separate it from} applying it to every language. (iii) Replacing the per-language re-examination thresholds with one
shared threshold, with the \camrev{replacement-candidate requirement} also lowered to
18{,}000 samples, gives a higher aggregate (0.953 on the held-out part) but
\camrev{the per-language F1 of 9 languages falls by more than 0.10 each\corrrev{; \cref{sec:calibration} states why the per-language thresholds are the default}.}

\paragraph{Transfer to a pretrained-vocabulary variant.} To test whether \corrrev{the
two corrections depend} on the dedicated tokenizer, we retrained the
\unilid-Mistral-Nemo variant with the recipe of \cref{sec:exp_setup} (\corrrev{a
training run independent of the one behind the \unilid-Mistral-Nemo row of
\cref{tab:lid_main}}) and re-derived every calibration component for the new
vocabulary by the procedures of this appendix: its own thresholds
\corrrev{$\delta_\lang$}, its own unseen-token treatment (\corrrev{509 languages whose trained
unseen-token values already lie at or below $c = -17$ are left unchanged}), and
its own high-entropy group, for which the criteria above select Banjar,
Scots, and Serbian in Latin script. \Cref{tab:calibrated_nemo} reports the
effect: \corrrev{the calibrated variant reaches} macro F1 \corrrev{0.950 against 0.912} on
the full pool, and the paired bootstrap interval of the improvement on the
held-out subset is \corrrev{$+0.0489$} with 95\% interval \corrrev{$[+0.0424, +0.0555]$}. The gain
is larger than for the dedicated-tokenizer model (\corrrev{$+0.039$} against \corrrev{$+0.024$}
in full-pool macro F1).

\begin{table}[htbp]
\centering
\small
\adjustbox{max width=\linewidth}{
\begin{tabular}{lrrr}
\toprule
& \multicolumn{2}{c}{\textbf{Full pool}} & \textbf{Held-out} \\
\cmidrule(lr){2-3} \cmidrule(lr){4-4}
\textbf{Configuration} & \textbf{Macro F1} & \corrrev{\textbf{Macro FPR ($\times 10^{5}$)}} & \textbf{Macro F1} \\
\midrule
\unilid-Mistral-Nemo (retrained) & \corrrev{0.912} & \corrrev{1.86} & \corrrev{0.895} \\
\quad + unseen-token constant & \corrrev{0.935} & \corrrev{1.79} & \corrrev{0.923} \\
\quad + re-examination (calibrated) & \corrrev{0.950} & \corrrev{1.62} & \corrrev{0.944} \\
\bottomrule
\end{tabular}}
\caption{\corrrev{The two corrections of \cref{sec:calibration}} applied to an independently retrained
\unilid-Mistral-Nemo variant, on the 45{,}377{,}279-line GlotLID-C pool
(full pool) and the 27{,}002{,}441-line held-out subset. FPR values are
multiplied by $10^5$. \corrrev{The retrained baseline is within $0.002$ macro F1 of
the \unilid-Mistral-Nemo row of \cref{tab:lid_main}.}}
\label{tab:calibrated_nemo}
\end{table}
 
\paragraph{Remaining errors.} \corrrev{Under calibrated \unilid,} \corrrev{930{,}576} wrong predictions
remain on the held-out part; \corrrev{99.1\%} of them are lines whose true language has
at least 18{,}000 training samples, and 88.6\% of those are confusions with
another such language, concentrated in close pairs (Indonesian and Standard
Malay, \corrrev{31{,}105} lines; Standard Arabic and Najdi Arabic; Mandarin and Wu
Chinese). No unseen-token treatment changes such a pair's score comparison
materially, so this is \corrrev{the limit of this class of corrections}; \cref{sec:disc}
sketches what reaching past it would require.

\end{document}